\documentclass[english]{article}
\usepackage[T1]{fontenc}
\usepackage[latin9]{inputenc}
\usepackage{amsmath}
\usepackage{amssymb}

\makeatletter

\usepackage{microtype}
\usepackage{graphicx}
\usepackage{subfigure}
\usepackage{booktabs} %
\usepackage{mathabx}

\usepackage{tikz}
\usepackage{pgfplots}

\usepackage{hyperref}

\usepackage[accepted]{icml2020}

\icmltitlerunning{Continuous Time Bayesian Networks with Clocks}

\makeatother

\usepackage{babel}
\begin{document}
\twocolumn[ \icmltitle{Continuous-Time Bayesian Networks with Clocks}
\begin{icmlauthorlist} \icmlauthor{Nicolai Engelmann}{tu-e} 
\icmlauthor{Dominik Linzner}{tu-e} 
\icmlauthor{Heinz Koeppl}{tu-e,tu-b} 
\end{icmlauthorlist}
\icmlaffiliation{tu-e}{Department of Electrical Engineering and Information Technology, Technische Universitat Darmstadt, Darmstadt, Germany}
\icmlaffiliation{tu-b}{Department of Biology, Technische Universitat Darmstadt, Darmstadt, Germany}
\icmlcorrespondingauthor{Nicolai Engelmann}{nicolai.engelmann@bcs.tu-darmstadt.de}
\vskip 0.3in ]

\printAffiliationsAndNotice{}

\begin{abstract}
Structured stochastic processes evolving in continuous time present a widely adopted framework to model phenomena occurring in nature and engineering. However, such models are often chosen to satisfy the Markov property to maintain tractability. One of the more popular of such memoryless models are Continuous Time Bayesian Networks (CTBNs). In this work, we lift its restriction to exponential survival times to arbitrary distributions. Current extensions achieve this via auxiliary states, which hinder tractability. To avoid that, we introduce a set of node-wise clocks to construct a collection of graph-coupled semi-Markov chains. We provide algorithms for parameter and structure inference, which make use of local dependencies and conduct experiments on synthetic data and a data-set generated through a benchmark tool for gene regulatory networks. In doing so, we point out advantages compared to current CTBN extensions.
\end{abstract}

\section{Introduction} \label{introduction}
Dynamics on networks are abundant. Examples span from small scales as in gene regulatory networks \cite{acerbi} to large scales as in population or opinion dynamics on social networks \cite{De2016}. All these processes depend on some form of memory. Genes interact via expression of molecules with maturation times that are not exponentially distributed \cite{Klann2012}, neuron- and social dynamics depend on past events \cite{Pillow2008}. 

While models with non-Markovian behavior exist, e.g. Hawkes processes \cite{Linderman2016}, they either follow a fixed parametrization that hinders their expressiveness, or are defined in discrete time as Dynamic Bayesian Networks \cite{Dagum1992} and are thus heavily over parametrized if no natural time scale exists.
An expressive, but Markovian model in continuous time are Continuous Time Bayesian Networks (CTBNs) \cite{Nodelman1995}. There have been several attempts in extending CTBNs to model non-Markovian behavior \cite{Nodelman2005, ctbn_ph, hiddenctbns}. However, in these extensions, a CTBN has to be inflated with arbitrary many auxiliary states. This hinders the scalability of such models. 

In this manuscript, we consider a different extension based on CTBNs, where we augment each local node of the network with its own clock process. This clock keeps track of the history of the node and determines the survival time of the node's state. The dynamics of the clocks can be instantiated by arbitrary survival time distributions.

We derive augmented CTBNs with clocks as a novel generative model along with likelihoods for a complete observed process in section 2. In section 3, we show how the exact parameter and structure inference can be performed. Finally, in sections 4 and 5 we show not only tractable parameter and structure learning for two example parametric survival time distributions, but as well demonstrate at hand of controlled synthetic experiments that CTBNs can fail as structure classifiers if provided data is non-Markovian, while augmented CTBNs perform well.

\subsection{Continuous Time semi-Markov and Markov Chains}
\label{ctsmcs}

We begin by laying out relevant key characteristics of homogeneous continuous time semi-Markov chains (CTSMC's) and explore their memoryless special case, continuous time Markov chains (CTMC's). A CTSMC is a non-Markovian jump processes with a renewal property evolving over continuous time, taking values in a countable state space $\mathcal{X}$. It can be completely described by a set of transition rates $\lambda: \mathcal{X}\times\mathbb{R}_{\geq0}\times\mathbb{R}_{\geq0}\times\mathcal{X}\rightarrow \mathbb{R}_{\geq0}$, where $\lambda(x,\tau,t;x')$ denotes the instantaneous rate of the process transitioning from state $x$ to $x'$ at time $t$ if its last transition happened at time $t-\tau$. Using them, we define the exit rate $\lambda(x,\tau,t)\equiv \sum_{x'\neq x}\lambda(x,\tau,t;x')$. A CTSMC is homogeneous if $\lambda(x,\tau,t;x')=\lambda(x,\tau;x')$ is dependent of $t$ only via $\tau$. In this work, we solely consider homogeneous CTSMC's in this sense. To generate a realization of the chain $\left\{ X(t)\right\}_{t\geq0}$, we have to first determine the time $s$ from arrival in state $x$ to the transition to another state, i.e. the survival time, from the cumulative distribution \cite{limnios}

\begin{align}\label{eq:survival}
F\left(s\,\mid\,x\right) &\equiv P\left(\text{"survival time in state $x$"}\leq s\right)\\
&= 1 - \exp\left(-\int_0^{s}\,\mathrm{d}\tau\; \lambda(x,\,\tau)\right)\nonumber,
\end{align}

with the survival function $\Lambda\left(s\,\mid\,x\right)\equiv 1-F\left(s\,\mid\,x\right)$. We then draw the next state $x'$ conditioned on the last state $x$ and the elapsed survival time $s$ from a categorical distribution $x'\mid x,s\sim \mathrm{Cat}(\lambda(x,s;x')/\lambda(x,s))$, where $x'\neq x$, $x,x'\in\mathcal{X}$. The property that the next state $x'$ is drawn independently of the time $s$ can be called time-direction independence \cite{Maes2009}. In this case, we define transition probabilities $\theta(x',x)\equiv\lambda(x,s;x')/\lambda(x,s)$ from state $x$ to $x'$. The quantities $\lambda(x,s)$ and $\theta(x)\equiv\{\theta(x',x)\mid x'\neq x\}$ then characterize the process. Since we need to know the time of the last transition to draw the survival time and next state, the process in general is non-Markovian. However, we can augment the chain $\left\{ X(t)\right\}_{t\geq0}$ by a process $\left\{ T(t)\right\}_{t\geq0}$, for which $T(t)=\tau$ always keeps track of the time since the last transition, resetting back to $0$ at each state change. We call $\left\{ T(t)\right\}_{t\geq0}$ the clock of the CTSMC. The two-component process $\left\{ X(t),\,T(t)\right\} _{t\geq0}$ on the product state space $\mathcal{X}\times \mathbb{R}_{\geq0}$ is then homogeneous and Markovian \cite{limnios}.

An important special case of a CTSMC is called a CTMC iff $\lambda(x,\tau;x')=\lambda(x;x')$ is constant for all $\tau$. This results in the process $\left\{ X(t)\right\}_{t\geq0}$ becoming Markovian and memoryless on its own, independent of $\left\{ T(t)\right\}_{t\geq0}$. As a result $F\left(s\,\mid\,x\right) = 1 - \exp\left(-\lambda(x)s\right)$ becomes fixed to be an exponential distribution. Note, that a CTMC always satisfies time-direction independence.

\subsection{Continuous Time Bayesian Networks}
A CTBN is a collection of $N$ homogeneous CTMCs $\left\{ X_n(t)\mid {t\geq0},n\in \left\{1,\dots,N\right\}\right\}$ evolving in parallel over a product state space $\mathcal{X}=\mathcal{X}_{1}\,\times\,\mathcal{X}_{2}\,\times\,\dots\,\times\,\mathcal{X}_{N}$ \cite{Nodelman1995}. As a consequence, we can
characterize the state of the CTBN $\boldsymbol{x}\in\mathcal{X}$ to be a collection of the states
of its constituting CTMCs  $\boldsymbol{x}=\left(x_{1},\,x_{2},\,\dots,\,x_{N}\right)$.
The processes of a CTBN are conditional on each other, with a dependency structure encoded in a directed graph $\mathcal{G}=\left(V,\,E\right)$ (which may contain cycles). Specifically, we associate vertices
$v_{n}\in V$ with processes $X_{n}$ and edges
$\left(v_{m},\,v_{n}\right)\in E$ with $X_{m}$ conditioning
$X_{n}$. We define the parent set of the $n$'th node by $\mathrm{par}_n\equiv\left\{v_m\mid \left(v_{m},\,v_{n}\right)\in E\right\}$. We can then collect the processes of parenting nodes as $U_n(t)\equiv\left\{X_m(t)\mid m\in\mathrm{par}_n \right\}$ taking values in $\mathcal{U}_n\equiv\bigtimes_{m\in\mathrm{par}_n }\mathcal{X}_m$. We denote realizations of this (vector-valued) process by $\boldsymbol{u}_{n}\in \mathcal{U}_n$.

 CTBNs are defined using the asynchronity of transitions in continuous time, which is that the probability of more than one jump at a given time instance scales in $o(h)$. Under this knowledge we can generate realizations of the CTBN by using the local quantities, node-wise conditional rates $\lambda_{n}:\mathcal{X}_n\times\mathcal{U}_n\rightarrow \mathbb{R}_{\geq 0}$ and node-wise conditional transition probabilities $\theta_{n}:\mathcal{X}_n\times\mathcal{X}_n\times\mathcal{U}_n\rightarrow [0,1]$. 
 Further, we can identify the global transition rate as the sum of node-wise rates $\lambda\left(\boldsymbol{x}\right)=\sum_{n=1}^{N}\lambda_n\left(x_{n},\,\boldsymbol{u}_{n}\right)$. Thus, the time until the global state change $\boldsymbol{x}\rightarrow\boldsymbol{x}'$ takes place can be again determined via \eqref{eq:survival} in the form of an exponential distribution for a CTMC.
 We can then determine the next state $\boldsymbol{x}'$ given $\boldsymbol{x}$ by subsequently drawing the changing process $X_{n}$ and its next state $x_{n}'$ from categorical distributions with mass functions
 
\begin{flalign}
\label{eq:ctbnjp}
&p\left(n\,|\,\boldsymbol{x}\right) =\frac{\lambda_n\left(x_{n},\,\boldsymbol{u}_{n}\right)}{\sum_{m}\lambda_m\left(x_{m},\,\boldsymbol{u}_{m}\right)}\\
&p\left(x_n'\,|\,\boldsymbol{x},n\right) =\frac{\lambda_n\left(x_{n},\,\boldsymbol{u}_{n};\,x_{n}'\right)}{\lambda_n\left(x_{n},\,\boldsymbol{u}_{n}\right)}.\nonumber
\end{flalign}

\section{Bayesian Networks of Generic CTSMC's}
\label{semictbns}

\subsection{Model Definiton}\label{model}

We define the augmented CTBN 
on a semi-continuous state space $\mathcal{X}_{1}\times\mathbb{R}_{\geq0}\times\mathcal{X}_{2}\times\mathbb{R}_{\geq0}\times\dots\times\mathcal{X}_{N}\times\mathbb{R}_{\geq0}$,
with $N$ local state variables $X_{n}\left(t\right)$ and local clocks
$T_{n}\left(t\right)$ for $n\in\left\{ 1,\,\dots,\,N\right\} $.
Like above, we express this collection in a vector $\left(\boldsymbol{x},\,\boldsymbol{\tau}\right)=\left(x_{1},\,x_{2},\,\dots,\,x_{N},\,\tau_{1},\,\tau_{2},\,\dots,\,\tau_{N}\right)$. We refer to the random variable associated with the collection of all local state variables by $\boldsymbol{X}\left(t\right)$ and the one associated with all local clocks by $\boldsymbol{T}\left(t\right)$.
Like in original CTBN's, we can argue, that under knowledge of all
(generalized) states, the global process obeys the Markov property. We fully inherit the semantics of a graph inducing the dependencies
from CTBN's and allow a conditioning only via the discrete state components in $\boldsymbol{X}\left(t\right)$ like
in other proposed CTBN extensions \cite{Nodelman2005, ctbn_ph, hiddenctbns}. Thus, a process being conditionally dependent on another process in the network, is only dependent on
its discrete state component and not its clock.

\begin{figure*}
    \centering
    \includegraphics[width=0.8\textwidth]{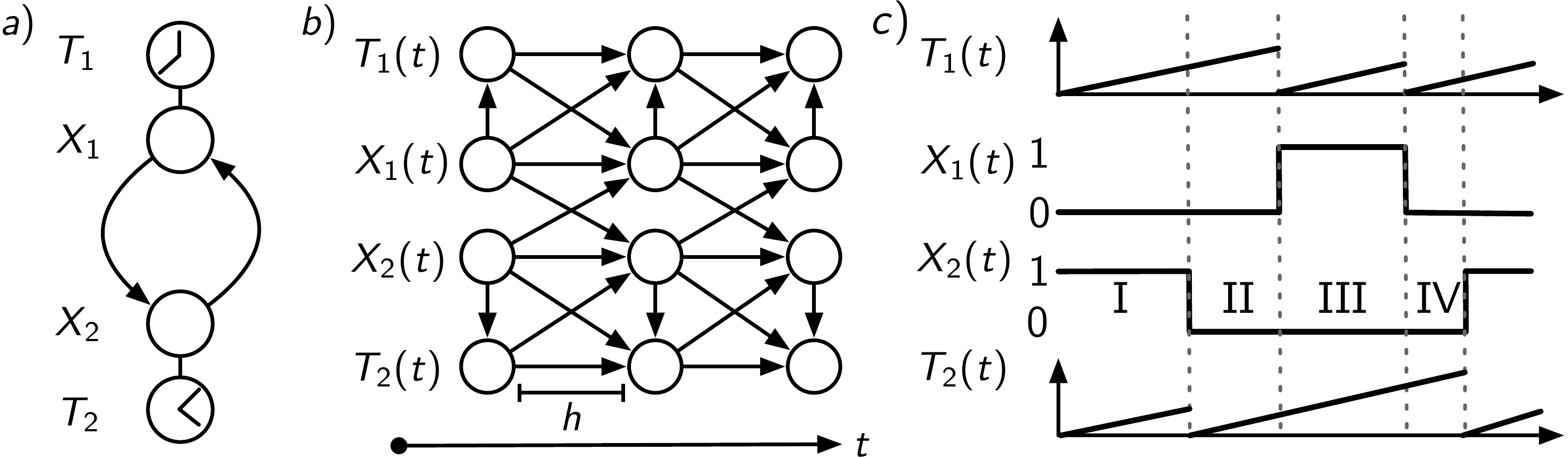}
    \caption{A graphic illustrating the semantic construction of the augmented CTBN. a) shows $N=2$ nodes with their local clocks. b) depicts their conditional relationships in a small time window $h$ as an equivalent Dynamic Bayesian Network. c) shows a sample trajectory for binary $X_1$ and $X_2$ highlighting the four classes of time-windows for $X_2$: full (I), censored (II), truncated and censored (III), and truncated (IV).}
    \label{fig:semantics}
\end{figure*}

We introduce local conditional transition rates $\lambda\left(x_{n},\,\tau_{n},\,\boldsymbol{u}_{n};\,x_{n}'\right)$
for a process $X_{n}\left(t\right)$ depending on its state $X_{n}\left(t\right)=x_{n}$,
its clock $T_{n}\left(t\right)=\tau_{n}$ and the states of other
processes $U_{n}\left(t\right)=\boldsymbol{u}_{n}$ referred to as
parent state like above. These rates are completely determined by our choice of the survival time distributions. We show this in Appendix A.3. Like above, we can define local exit rates $\lambda\left(x_{n},\,\tau_{n},\,\boldsymbol{u}_{n}\right)=\sum_{x_n'\neq x_n}\lambda\left(x_{n},\,\tau_{n},\,\boldsymbol{u}_{n};\,x_{n}'\right)$.
Similarly to CTBN's, if the associated graph
$\mathcal{G}$ has an edge $\left(v_{m},\,v_{n}\right)$, then $X_{m}\left(t\right)\in U_{n}\left(t\right)$
is a parent of $X_{n}\left(t\right)$. We also show in Appendix A.3, that
using the Markov property of the global process $\left\{\boldsymbol{X}(t),\boldsymbol{T}(t)\right\}$, the asynchronous
transition condition and the inherited properties of the constituting
CTSMC's are sufficient to fully characterize the global survival time distribution
and transition probabilities in terms of the local conditional transition rates
of the processes. We do so by integrating the probabilities of all constituting processes keeping their current states $x_n$ over small time windows of duration $h$ and thus find the survival function of the global process with respect to $\boldsymbol{X}(t)$ in the limit $h \rightarrow 0$ by

\begin{align}
\label{eq:globalsurv}
\Lambda\left(s\,|\,\boldsymbol{x},\,\boldsymbol{\tau}\right) & =\exp\left(-\sum_{n=1}^{N}\int_{\tau_{n}}^{\tau_{n}+s}\,\mathrm{d}\sigma\,\lambda\left(x_{n},\,\sigma,\,\boldsymbol{u}_{n}\right)\right)\nonumber\\
&=\prod_{n=1}^{N}\frac{\Lambda_{n}\left(s+\tau_{n}\,|\,x_{n},\,\boldsymbol{u}_{n}\right)}{\Lambda_{n}\left(\tau_{n}\,|\,x_{n},\,\boldsymbol{u}_{n}\right)},
\end{align}

in terms of local survival functions $\Lambda_{n}\left(s\,|\,x_{n},\,\boldsymbol{u}_{n}\right)=\exp\left(-\int_{0}^{s}\,\mathrm{d}\sigma\,\lambda\left(x_{n},\,\sigma,\,\boldsymbol{u}_{n}\right)\right)$ conditioned on the parents' state $\boldsymbol{u}_{n}$. An illustration of this approach is given in fig. {\ref{fig:semantics} b)}. Consequently, introducing the global $F\left(s\,|\,\boldsymbol{x},\,\boldsymbol{\tau}\right) = 1-\Lambda\left(s\,|\,\boldsymbol{x},\,\boldsymbol{\tau}\right)$ and local $F_{n}\left(s\,|\,x_{n},\,\boldsymbol{u}_{n}\right) = 1-\Lambda_{n}\left(s\,|\,x_{n},\,\boldsymbol{u}_{n}\right)$ cumulative distribution functions along with their local density functions $f_{n}\left(s\,|\,x_{n},\,\boldsymbol{u}_{n}\right)$, we can give the density function of the global survival time by

\begin{align}
\label{eq:smctbnsurvdens}
 &p\left(s\,|\,\boldsymbol{x},\boldsymbol{\tau}\right) = \frac{\mathrm{d}F\left(s\,|\,\boldsymbol{x},\,\boldsymbol{\tau}\right)}{\mathrm{d}s}\\
 &\;=\sum_{n=1}^{N}\frac{f_{n}\left(s+\tau_{n}\,|\,x_{n},\,\boldsymbol{u}_{n}\right)}{\Lambda_{n}\left(\tau_{n}\,|\,x_{n},\,\boldsymbol{u}_{n}\right)}\prod_{k\neq n}\frac{\Lambda_{k}\left(s+\tau_{k}\,|\,x_{k},\,\boldsymbol{u}_{k}\right)}{\Lambda_{k}\left(\tau_{k}\,|\,x_{k},\,\boldsymbol{u}_{k}\right)}\nonumber.
\end{align}

Note, that we can describe $p\left(s\,|\,\boldsymbol{x},\boldsymbol{\tau}\right)$ in terms of only a finite set of local distributions, which are independent of the current clock values $\boldsymbol{\tau}$. The query for the next global state $\boldsymbol{x}'$ can be answered by subsequently drawing the changing process $X_n$ and its next state $x_n'$ given the clock values $\boldsymbol{\tau}$ and the duration $s$ from categorical distributions with mass functions

\begin{align}
\label{eq:smctbnjp}
&p\left(n\,|\,\boldsymbol{x},\boldsymbol{\tau},s\right) =\frac{\lambda_n\left(x_{n},\,\tau_{n}+s,\,\boldsymbol{u}_{n}\right)}{\sum_{m}\lambda_m\left(x_{m},\,\tau_{m}+s,\,\boldsymbol{u}_{m}\right)}\\
&p\left(x_n'\,|\,\boldsymbol{x},\boldsymbol{\tau},s,n\right) =\frac{\lambda_n\left(x_{n},\,\tau_{n}+s,\,\boldsymbol{u}_{n};\,x_{n}'\right)}{\lambda_n\left(x_{n},\,\tau_{n}+s,\,\boldsymbol{u}_{n}\right)}.\nonumber
\end{align}

As a result, we update the clock values $\tau_n \rightarrow 0$ for the changing process and $\forall m\neq n: \tau_m \rightarrow \tau_m + s$ for the others. We provide a detailed derivation of the global survival time density and the transition probabilities in Appendix A.3. We are now in the position to generate CTBNs with any parametrized distribution for any state $x_n$ and condition $\boldsymbol{u}_n$ for all $n$. It is no coincidence, that \eqref{eq:smctbnsurvdens} resembles the density of a minimum distribution, which renders trajectory sampling straight forward and efficient for parametric local survival time distributions despite having to evaluate truncated distributions. This can be done with a Gillespie algorithm (outlined in Appendix C), subsequently drawing global survival times, changing processes and next states. After each step, we have to update $\boldsymbol{x}$ and $\boldsymbol{\tau}$ accordingly. Note, that we share the occurrence of truncations with generalized semi-Markov chains, in which triggered events also ``interrupt''
the survival procedure and we have to ``continue'' bearing truncated
distributions \cite{cassandras}.

\subsubsection{Parametrization of Local Survival Time Distributions}
\label{parlocal}

In \eqref{eq:survival} from section \ref{ctsmcs} we can see, that the exit rates $\lambda(x,s)$ can be determined if $\Lambda\left(s\,|\,x\right)$ is known. This also holds for the local $\lambda_n\left(x_{n},\,s,\,\boldsymbol{u}_{n}\right)$ given $\Lambda_{n}\left(s\,|\,x_{n},\,\boldsymbol{u}_{n}\right)$ in the augmented CTBN. Despite being able to parametrize local distributions by transition rate functions directly, many distributions are conveniently parametrized by a different set of quantities. In consideration of possible applications, we therefore introduce hyperparameters $\boldsymbol{\phi}\equiv\left\{\forall n,x_n,\boldsymbol{u}_n:\,\phi_n\left(x_n,\,\boldsymbol{u}_n\right)\right\}$ representing a chosen distribution per process $n$, its state $x_n$ and parent state $\boldsymbol{u}_n$, and consequently let $\lambda_n\left(x_{n},\,s,\,\boldsymbol{u}_{n}\right)$ and $\Lambda_{n}\left(s\,|\,x_{n},\,\boldsymbol{u}_{n}\right)$ take on a functional form dependent on $\phi_n\left(x_n,\,\boldsymbol{u}_n\right)$. We also restrict our considerations to cases, where the local processes in the augmented CTBN obey time-direction independence between state changes, which allows the factorization $\lambda_{n}\left(x_{n},\,\tau_{n},\,\boldsymbol{u}_{n};\,x_{n}'\right) \equiv \theta_n\left(x_{n}'\mid\,x_n,\,\boldsymbol{u}_{n}\right)\lambda_{n}\left(x_{n},\,\tau_{n},\,\boldsymbol{u}_{n}\right)$. As a consequence, $\boldsymbol{\phi}$ suffices to describe all survival time distributions. To account for this, we write $\lambda_n\left(x_{n},\,s,\,\boldsymbol{u}_{n}\right)\equiv \lambda_n\left(s,\,\phi\left(x_n,\,\boldsymbol{u}_n\right)\right)$ and $\Lambda_{n}\left(s\,|\,x_{n},\,\boldsymbol{u}_{n}\right)\equiv \Lambda_{n}\left(s\,|\,\phi\left(x_n,\,\boldsymbol{u}_n\right)\right)$. To give a short example, for a Weibull distribution in shape-rate parametrization ($k$, $b$), we would have the free parameters $\phi_n\left(x_n,\,\boldsymbol{u}_n\right)\equiv \left(k_{n}\left(x_n,\,\boldsymbol{u}_n\right), b_{n}\left(x_n,\,\boldsymbol{u}_n\right)\right)$ for local state $x_n$ and parent state $\boldsymbol{u}_n$, giving

\begin{align}\label{eq:exampleweibull}
&\Lambda_{n}\left(s\,|\,\phi_{n}\left(x_n,\,\boldsymbol{u}_n\right)\right)\\
&\qquad\qquad = \exp\left(-b_{n}\left(x_n,\,\boldsymbol{u}_n\right)s^{k_{n}\left(x_n,\,\boldsymbol{u}_n\right)}\right)\nonumber\\
&\lambda_n\left(s,\,\phi_{n}\left(x_n,\,\boldsymbol{u}_n\right)\right)\nonumber\\
&\qquad\qquad = b_{n}\left(x_n,\,\boldsymbol{u}_n\right)k_{n}\left(x_n,\,\boldsymbol{u}_n\right)s^{k_{n}\left(x_n,\,\boldsymbol{u}_n\right) - 1}.\nonumber
\end{align}

\subsection{Path Measure}

The global path measure of our augmented CTBN can be derived from the expressions in \eqref{eq:smctbnsurvdens} and \eqref{eq:smctbnjp}, and factorizes in terms of its discrete $\boldsymbol{X}\left(t\right)$ and continuous $\boldsymbol{T}\left(t\right)$ components for single transitions as a consequence of our simplifications in section \ref{parlocal}. For a fixed number $M$ of transitions it is given by a product of $M$ global density function
evaluations \eqref{eq:smctbnsurvdens} and $M$ evaluations of the mass functions representing the categorical transition directions \eqref{eq:smctbnjp}. Shown in detail in Appendix A.4, we obtain the density for a single time-window from entry in state $\boldsymbol{x}$ with clock configuration $\boldsymbol{\tau}$ on to our transition to $\boldsymbol{x}'$ after duration $s$ with the $n$-th process changing by

\begin{flalign}
\label{eq:lhlocal}
 &p\left(\boldsymbol{x}',\,\boldsymbol{\tau}'\,|\,\boldsymbol{x},\,\boldsymbol{\tau},\,\boldsymbol{\theta},\,\boldsymbol{\phi}\right)
 = p\left(\boldsymbol{x}'\,|\,\boldsymbol{x},\,\boldsymbol{\theta}\right)
 p\left(\boldsymbol{\tau}'\,|\,\boldsymbol{x}',\,\boldsymbol{x},\,\boldsymbol{\tau},\,\boldsymbol{\phi}\right)\nonumber\\
 &\qquad\quad= \theta_n\left(x_{n}'\,|\,x_{n},\,\boldsymbol{u}_{n}\right)\frac{f_n\left(s+\tau_{n}\,|\,\phi_n\left(x_{n},\,\boldsymbol{u}_{n}\right)\right)}{\Lambda_{n}\left(\tau_{n}\,|\,\phi_n\left(x_{n},\,\boldsymbol{u}_{n}\right)\right)}\nonumber\\
 &\qquad\qquad \times \prod_{k\neq n}\frac{\Lambda_{k}\left(s+\tau_{k}\,|\,\phi_k\left(x_{k},\,\boldsymbol{u}_{k}\right)\right)}{\Lambda_{k}\left(\tau_{k}\,|\,\phi_k\left(x_{k},\,\boldsymbol{u}_{k}\right)\right)},
\end{flalign}

while the factor $p\left(\boldsymbol{x}'\,|\,\boldsymbol{x},\,\boldsymbol{\theta}\right)$ is independent of the clocks and the survival times. We recognize this from embedded Markov chains in the continuous time context \cite{pinsky}. Further, we must specify

\begin{align*}
\boldsymbol{x}'&=\left(x_{1},\,\dots,\,x_{n-1},\,x_{n}',\,x_{n+1},\,\dots,\,x_{N}\right)\\
\boldsymbol{\tau}'&=\left(\tau_{1}+s,\,\dots,\,\tau_{n-1}+s,\,0,\,\tau_{n+1}+s,\,\dots,\,\tau_{N}+s\right).
\end{align*}

Since successive time-windows condition on $\left(\boldsymbol{x}',\,\boldsymbol{\tau}'\right)$
as the new ``starting point'', we can construct the measure for $M$ transitions by a product of $M$ such time-windows. It is also reasonable to assume the final window to be fully censored, so that there is no transition exactly at the end. In this case, the measure for this time-window reduces
to an evaluation of the global survival function \eqref{eq:globalsurv}. Consequently, gathering $\mathbf{X}=\left\{ \boldsymbol{x}^{(0)},\,\boldsymbol{x}^{(1)},\,\dots,\,\boldsymbol{x}^{(M)}\right\} $
and $\mathbf{T}=\left\{ \boldsymbol{\tau}^{(0)},\,\boldsymbol{\tau}^{(1)},\,\dots,\,\boldsymbol{\tau}^{(M)}\right\}$, we can write

\begin{align}
\label{eq:lhglobal}
&p\left(\mathbf{X},\,\mathbf{T}\,|\,\boldsymbol{\theta},\,\boldsymbol{\phi}\right)
= p\left(\mathbf{X}\,|\,\boldsymbol{\theta}\right)p\left(\mathbf{T}\,|\,\mathbf{X},\,\boldsymbol{\phi}\right)\\
&\qquad= \prod_{m=0}^{M-1}p\left(\boldsymbol{x}^{(m+1)}\,|\,\boldsymbol{x}^{(m)},\,\boldsymbol{\theta}\right)\nonumber\\
 &\qquad\quad \times \prod_{m=0}^{M-1}p\left(\boldsymbol{\tau}^{(m+1)}\,|\,\boldsymbol{x}^{(m+1)},\,\boldsymbol{x}^{(m)},\,\boldsymbol{\tau}^{(m)},\,\boldsymbol{\phi}\right).\nonumber
\end{align}

which can be made explicit by directly inserting the respective terms outlined in \eqref{eq:lhlocal}. We give the factors of the path measure associated with the survival times for Weibull and Gamma distributions in section \ref{examples} and Appendix A.5. Under knowledge of \eqref{eq:lhglobal}, we can
work on the question of inference of model parameters of the augmented
CTBN. Before we do so, however, we want to briefly address related work.

\subsection{Related Work}

In the following we want to argue about the design choices of our model and compare them to models existing in literature. This is a discussion about different clock augmentations. Proposed extensions to CTBNs using phase-type distributions
\cite{ctbn_ph}\cite{Nodelman2005} and hidden nodes \cite{hiddenctbns} allow per-process survival time related structures. In both models, multiple
so-called phases keep track of the survival time and are associated
to a so-called generalized state and only this generalized state
is relevant for the conditions imposed on a dependent process. The CTBN extension based on hidden nodes in addition only allows conditioning of nodes on binary state spaces. This limits the models applicability and requires comparably large networks to model nodes with many states and arbitrary non-exponential survival time distributions.

While phase-type CTBNs are similar in terms of expressiveness, they suffer from two distinct problems which we aim to avoid. The first problem is limited tractability due to the introduction of auxiliary states, the second being the heavy over-parametrization of phase-type distributions \cite{Hobolth2019}. We alleviate both problems by replacing arbitrary auxiliary states with, also arbitrary, survival time distributions with few parameters.

\section{Inference of Model Parameters}

Model inference can be done with varying efficiency, depending on the choice of the survival time distribution. We can see in \eqref{eq:lhglobal}, that the likelihood factors not only in single time-window terms, but also in transition probabilities $\boldsymbol{\theta}$ and survival time distribution parameters $\boldsymbol{\phi}$. We can therefore give the posterior for independent $\boldsymbol{\theta}$ and $\boldsymbol{\phi}$ by

\begin{align}\label{eq:postpars}
p\left(\boldsymbol{\theta},\,\boldsymbol{\phi}\,|\,\mathbf{X},\,\mathbf{T}\right)\;
 & \propto \; p\left(\mathbf{X},\,\mathbf{T}\,|\,\boldsymbol{\theta},\,\boldsymbol{\phi}\right)p\left(\boldsymbol{\theta}\right)p\left(\boldsymbol{\phi}\right)\nonumber\\
 &= p\left(\mathbf{X}\,|\,\boldsymbol{\theta}\right)p\left(\boldsymbol{\theta}\right)p\left(\mathbf{T}\,|\,\mathbf{X},\,\boldsymbol{\phi}\right)p\left(\boldsymbol{\phi}\right)\nonumber\\
 & \propto \; p\left(\boldsymbol{\theta}\,|\,\mathbf{X}\right)p\left(\boldsymbol{\phi}\,|\,\mathbf{X},\,\mathbf{T}\right).
\end{align}

As a consequence, we can infer $\boldsymbol{\theta}$ traditionally in terms of transition probabilities of an embedded Markov chain \cite{pinsky}. This is well studied, and can be done analytically using a conjugate Dirichlet prior, so we refer to the original literature \cite{Nodelman2003} for the expressions. In \eqref{eq:lhlocal}, we can also see, that each single time-window only depends on parameters $\theta_n\left(x_{n}'\,|\,x_{n},\,\boldsymbol{u}_{n}\right)$ and $\phi_n\left(x_{n},\,\boldsymbol{u}_{n}\right)$ associated with the relevant local $x_{n}$ and parent state $\boldsymbol{u}_{n}$ configurations. If we assume all $\phi_n\left(x_{n},\,\boldsymbol{u}_{n}\right)$ to be mutually independent, $p\left(\boldsymbol{\phi}\,|\,\mathbf{X},\,\mathbf{T}\right)$ also factors in the local marginal posteriors $p\left(\phi_{n}\left(x_{n},\,\boldsymbol{u}_{n}\right)\,|\,\mathbf{X},\,\mathbf{T}\right)$, significantly reducing the curse of dimensionality in associated inference and learning problems. In this case, the marginal posterior $p\left(\phi_{n}\left(x_{n},\,\boldsymbol{u}_{n}\right)\,|\,\mathbf{X},\,\mathbf{T}\right)$ can be determined via

\begin{align}\label{eq:postsjsingle}
&p\left(\phi_{n}\left(x_{n},\,\boldsymbol{u}_{n}\right)\,|\,\mathbf{X},\,\mathbf{T}\right)\;\\
&\quad\propto \; \prod^M_{m=1} \frac{f_n\left(s^{(m)}+\tau^{(m)}_{n}\,\Big|\,\phi_n\left(x_{n},\,\boldsymbol{u}_{n}\right)\right)^{1-c\left(m,\,n,\,x_{n},\,\boldsymbol{u}_{n}\right)}}{\Lambda_{n}\left(\tau^{(m)}_{n}\,\Big|\,\phi_n\left(x_{n},\,\boldsymbol{u}_{n}\right)\right)}\nonumber\\
\nonumber&\qquad \times \Lambda_{n}\left(s^{(m)}+\tau^{(m)}_{n}\,\Big|\,\phi_n\left(x_{n},\,\boldsymbol{u}_{n}\right)\right)^{c\left(m,\,n,\,x_{n},\,\boldsymbol{u}_{n}\right)}\\
&\qquad \times p\left(\phi_{n}\left(x_{n},\,\boldsymbol{u}_{n}\right)\right)\nonumber,
\end{align}

where $s^{(m)}=\max\left\{\boldsymbol{\tau}^{(m+1)}-\boldsymbol{\tau}^{(m)}\right\}$ is the duration of the $m$-th time-window and $c\left(m,\,n,\,x_{n},\,\boldsymbol{u}_{n}\right)=1$ if it did not end with a transition of $X_n$, i.e. it is censored, otherwise $0$. Note, that if $\tau^{(m)}_{n}=0$, the denominator is $1$, hence the time-window is not truncated for $X_n$. An illustration of the four possible cases of censoring and truncation can be seen in fig. \ref{fig:semantics} c). If we omit the prior on the rhs of \eqref{eq:postsjsingle}, we obtain the factor of the likelihood for $\phi_{n}\left(x_{n},\,\boldsymbol{u}_{n}\right)$. By inserting our example \eqref{eq:exampleweibull} from section \ref{parlocal} and recognizing $f_n\left(s\,|\,\phi_{n}\left(x_n,\,\boldsymbol{u}_n\right)\right) = \lambda_n\left(s,\,\phi_{n}\left(x_n,\,\boldsymbol{u}_n\right)\right)\Lambda_n\left(s\,|\,\phi_{n}\left(x_n,\,\boldsymbol{u}_n\right)\right)$, we immediately obtain the expression for the Weibull case.

If we assume not a fixed graph $\mathcal{G}$ imposing structural dependencies in the augmented CTBN but a distribution over graphs $p\left(\mathcal{G}\right)$, we render the parent sets $\mathrm{par}_n$ (and the constitution of $U_n$) and consequently the $\boldsymbol{\theta}$ and $\boldsymbol{\phi}$ dependent of $\mathcal{G}$, and can find the posterior of $\mathcal{G}$ by

\begin{align}\label{eq:postgraph}
&p\left(\mathcal{G}\,|\,\mathbf{X},\,\mathbf{T}\right)
 \;\propto\; p\left(\mathbf{X}\,|\,\mathcal{G}\right) p\left(\mathbf{T}\,|\,\mathbf{X},\,\mathcal{G}\right) p\left(\mathcal{G}\right)\\
 \nonumber&=\left(\int_{\boldsymbol{\Theta}}\,\mathrm{d}\boldsymbol{\theta}\,p\left(\mathbf{X}\,|\,\boldsymbol{\theta}\right)p\left(\boldsymbol{\theta}\,|\,\mathcal{G}\right)\right.\\
 &\quad\left.\times\int_{\boldsymbol{\Phi}}\,\mathrm{d}\boldsymbol{\phi}\,p\left(\mathbf{T}\,|\,\mathbf{X},\,\boldsymbol{\phi}\right)p\left(\boldsymbol{\phi}\,|\,\mathcal{G}\right) \right)p\left(\mathcal{G}\right)\nonumber,
\end{align}

where again, while the leftmost integral is analytically solvable, the rightmost can be reduced to a product of low-dimensional integrations under the mentioned assumption that the $\phi_n\left(x_{n},\,\boldsymbol{u}_{n}\right)$ are independent. Then, the term becomes a product of integrals over the single $\phi_n\left(x_{n},\,\boldsymbol{u}_{n}\right)$ with the integrals containing the likelihood term from the rhs of \eqref{eq:postsjsingle}. In addition to that, our approach preserves structure modularity. Consequently, as long as the prior $p\left(\mathcal{G}\right)=\prod_{n=1}^{N}p\left(\mathcal{G}_{n}\right)$ factors in the same way, where $\mathcal{G}_{n}=\left(V,\,E_{n}\right)$ consists of only the incoming edges to process $X_n$, we can calculate the posterior of the full graph via $p\left(\mathbf{X},\,\mathbf{T}\,|\,\mathcal{G}\right)p\left(\mathcal{G}\right)=\prod_{n=1}^{N}p\left(\mathbf{X},\,\mathbf{T}\,|\,\mathcal{G}_{n}\right)p\left(\mathcal{G}_{n}\right)$ and then $\mathcal{G}=\left(V,\,\bigcup_{n}E_{n}\right)$, even more reducing dimensionality.

Because the augmented CTBN preserves the properties layed out in \eqref{eq:postpars} and \eqref{eq:postgraph}, it is - in contrast to existing CTBN extensions - possible to perform exact inference fully analytical for some non-exponential distributions. In Appendix A.6 we give the analytical marginal likelihoods \eqref{eq:postgraph} and posteriors \eqref{eq:postpars} along with their conjugate priors and sufficient statistics for an augmented CTBN with Rayleigh survival time distributions as an example.

\section{Example Survival Time Distributions}\label{examples}

The description of the augmented CTBN given above is valid for a broad family of parametric and non-parametric survival time distributions. Before we conduct experiments, we want to give two specific implementations of the augmented CTBN based on Weibull and Gamma distributions. Those are not only important in survival analysis \cite{coxsurv} and queueing theory, but also exhibit distinct numerical properties. In addition to that, Weibull CTBNs have a non-exponential tail in the general case and are therefore naturally hard to approximate by PH distributions used in related publications. Gamma r.v.'s on the other hand can be built by sums of exponential r.v.'s in the case of integer shape parameters.

\subsection{Weibull CTBN's}

For the Weibull distribution, we chose the parametrization outlined in section \ref{parlocal} (see \eqref{eq:exampleweibull}. To keep our notation simple, we omit the $\left(x,\,\boldsymbol{u}\right)$ in the following, if no ambiguity occurs. Consequently, the factors of the likelihood associated with
only the survival time parameters $\phi_n\left(x_n,\,\boldsymbol{u}_n\right)\equiv \left(k_{n}, b_{n}\right)$
then read

\begin{align*}
 &p\left(k_{n},\,b_{n}\,|\,\mathbf{X},\,\mathbf{T}\right)\\
 &\quad\propto\; p\left(k_{n},\,b_{n}\right) b_{n}^{\left|S_{f}\right|}k_{n}^{\left|S_{f}\right|}\left(\prod_{m=1}^{\left|S_{f}\right|}s_{f,m}\right)^{k_{n}-1}\\
 &\quad\exp\left(-b_{n}\left(\sum_{m=1}^{\left|S_{f}\right|}s_{f,m}^{k_{n}}+\sum_{m=1}^{\left|S_{c}\right|}s_{c,m}^{k_{n}}-\sum_{m=1}^{\left|S_{t}\right|}s_{t,m}^{k_{n}}\right)\right),
\end{align*}

where the set $S_{f}=\left\{ s_{f,1},\,s_{f,2},\,\dots\right\} $
contains all values, the clock would have displayed just before the considered process transitioned, i.e. $\tau_n + s$. Furthermore, $S_{c}$ gathers all clock samples, at which a parent change happens, while $S_{t}$ gathers the samples at the beginning of the time-windows, i.e.
the truncation times. We outline the construction with focus on sample efficiency in Appendix A.5. An advantage of the Weibull distribution is
its exponential form and monomial rates. This allows us to evaluate
sums of monomials in a logarithmic domain directly, which results
in an improved numerical stability, especially compared with a Gamma
distribution. Another interesting property of the distribution is
its expressibility of sub- and superexponential tails.

\subsection{Gamma CTBN's}

The parameterization of the Gamma distributions in our CTBN is chosen
by a rate $\alpha$ and shape $\beta$ parameter similar to the Weibull
case. The local survival functions of the CTBN for the $n$-th process
can then be given by $\Lambda_{n}\left(s\,|\,\phi_{n}\left(x_{n},\,\boldsymbol{u}_{n}\right)\right)=\frac{\Gamma\left(\alpha_{n}\left(x,\,\boldsymbol{u}\right),\,\beta_{n}\left(x,\,\boldsymbol{u}\right)s\right)}{\Gamma\left(\alpha_{n}\left(x,\,\boldsymbol{u}\right)\right)}$,
where the $\Gamma$ in the numerator denotes the upper incomplete gamma function. We can
retrieve the exit rates via $\lambda_{n}\left(s,\,\phi_{n}\left(x_{n},\,\boldsymbol{u}_{n}\right)\right)\Gamma\left(\alpha_{n}\left(x,\,\boldsymbol{u}\right),\,\beta_{n}\left(x,\,\boldsymbol{u}\right)s\right)=\beta_{n}\left(x,\,\boldsymbol{u}\right)^{\alpha_{n}\left(x,\,\boldsymbol{u}\right)}s^{\alpha_{n}\left(x,\,\boldsymbol{u}\right)-1}\exp\left(-\beta_{n}\left(x,\,\boldsymbol{u}\right)s\right)$. The likelihood associated with the Gamma distribution can be found in Appendix A.5. Numerically disadvantageous, the logarithm of the incomplete gamma function is comparably hard to obtain and if evaluated
in the linear domain first, a wide range of parameter values for different
$\left(x,\,\boldsymbol{u}\right)$ can cause the likelihood to contain
ratios with numerators and denominators close to or below machine precision.

\section{Experimental Work}

In this section, we give simulations for parameter and structure inference based on synthetic trajectories from augmented CTBNs generated using the Gillespie sampling procedure mentioned in section \ref{model}. In the end, we highlight a result for structure inference of an augmented CTBN under a Gamma distribution assumption on trajectories obtained through GeneNetWeaver \cite{Schaffter2011} for a gene regulatory network. 

\begin{figure*}[t]
    \centering
    \includegraphics[width=\textwidth]{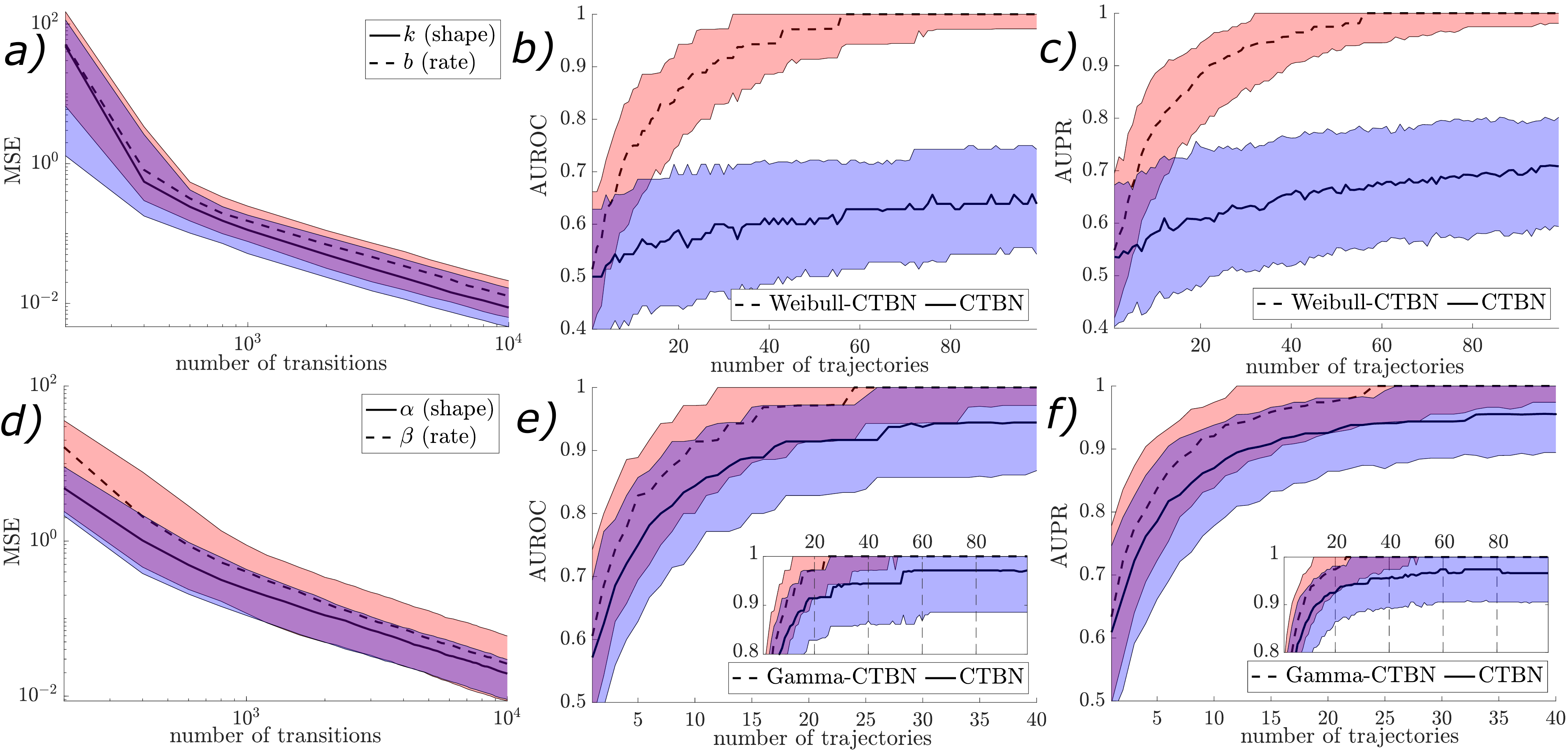}
    \caption{a) mean squared error (MSE) distribution for varying numbers of transitions in the Weibull parameter inference scenario. $80\%$ of the calculated MSE's are bounded by the colored area. The blue area corresponds to the shape and the red area to the rate. An exponential convergence to the true parameter sets can be seen as a straight line in the dual log plot. b) \& c) AUROC and AUPR for the Weibull structure inference scenario for varying numbers of trajectories compared with a traditional CTBN. With a mean shape of around $k\approx15$ for the generated Weibull distributions, the traditional CTBN slowly approaches the true graph, but fails to classify the ground truth with the given number of trajectories. d) MSE similar to a) for the Gamma parameter inference scenario. e) \& f) AUROC and AUPR for the Gamma structure inference scenario similar to the Weibull case. A generally faster convergence (mean shape around $\alpha\approx8$) compared to Weibull can be seen and although the traditional CTBN fails to classify the graph, it approaches it closely (see inlet plot).}
    \label{fig:inference}
\end{figure*}

\subsection{Parameter and Structure Inference on Synthetic Trajectories}

For this scenario, we sampled trajectories of varying lengths from random augmented CTBN's with four nodes via the Gillespie algorithm from Appendix C and performed parameter and structure inference on the trajectories. First, we describe, how the random graph including the random survival time distribution parameters is sampled. From a uniform degree distribution we determined the size $\left|\mathcal{U}_n\right|$ of the independent parent set for node $n$. Subsequently we drew a sample from a \emph{uniform} Dirichlet distribution to obtain a random categorical distribution. To determine the nodes in the parent set, we then chose the $\left|\mathcal{U}_n\right|$ maximum values in the obtained vector. We did so for each node to obtain random independent parent sets. Then, for each independent parameter set in the CTBN associated with the sampled graph, we sampled random parameters for the Weibull and Gamma distributions again from Gamma distributions. Their shapes and rates (Gamma: $\alpha_1=40, \beta_1=5$ for the shape and $\alpha_2=25, \beta_2=2.5$ for the rate generation, Weibull: $\alpha_1=8, \beta_1=0.5$ for the shape and $\alpha_2=5, \beta_2=3$ for the rate) were chosen to lead to reasonable model parameters with a significant variance in the samples and pronounced non-exponential appearance.

\begin{figure*}[t]
    \centering
    \includegraphics[width=\textwidth]{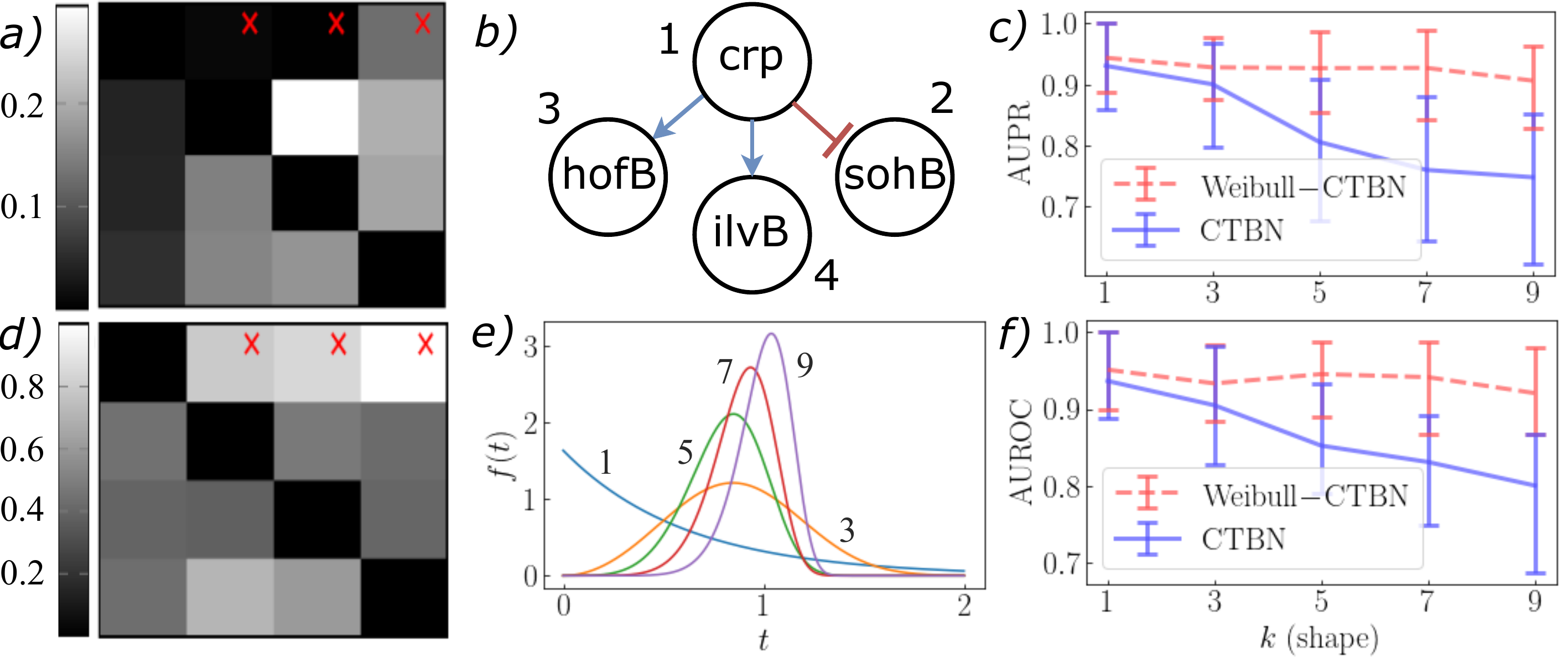}
    \caption{a) The weighted adjacency matrix for the gene regulatory network experiment obtained via structure inference from a traditional CTBN, while d) gives the result for the Gamma CTBN. The grayscales indicate the values of the posterior distribution. The traditional CTBN predicts wrong edges, while the Gamma CTBN creates an almost fully connected graph despite a Laplace prior on the edges. The high confidence edges, however, are located at the correct positions in the Gamma case. b) The GRN, from which the data has been generated. It is sourced from the gold standard for E. Coli stored in GeneNetWeaver. c) \& f) The AUROC scores over the chosen shape parameters for the Weibull distributions in the structure inference scenario with controlled shape parameters. e) An illustration of the five Weibull distributions with fixed shape. The number next to each curve corresonds to its shape parameter value $k$. The rate parameters stayed random for each graph, process, local state and parent state combination.}
    \label{fig:grn}
\end{figure*}

\subsubsection{Parameter Inference}

For parameter inference, we carry out an experiment for Weibull and Gamma CTBN's each. We give a plot of the mean squared error (MSE) showing quantiles over the $1\,000$ trajectories in fig. \ref{fig:inference}. For each of the CTBN's, we chose a random graph and sampled $1\,000$ trajectories with a size of $10\,000$ transitions each from it. Note, that the amount of transitions corresponds to the global process. We therefore obtained a significantly smaller amount of samples per parameter set $\phi_n\left(x_n,\,\boldsymbol{u}_n\right)$ of a single node, local state and parent state. Then, we inferred the parameters for each graph given different numbers of transitions reaching to the maximum of $10\,000$. For the prior, we chose a uniform prior imposing box constraints from $0.1$ to $100$ for both parameters. Since we could only perform numerical posterior updates on a discrete grid, we chose to always evaluate a single likelihood to obtain an arbitrary close approximation of the posterior mode in this experiment. To calculate the MAP-estimates, we employed a limited-memory BFGS algorithm using projected gradients to stay in the bounds dictated by the uniform prior. We did so for each of the $1\,000$ trajectories and then calculated the MSE of each parameter estimate for each inferred number of time-windows over all trajectories. An exponential convergence to the true parameter values can be observed in the plots in fig. \ref{fig:inference}.

\subsubsection{Structure Inference}\label{structureinference}

We performed the structure inference experiment for Gamma and Weibull CTBN's separately as well and compared the results to traditional CTBN's. For each, we generated $500$ different random graphs. The following procedure was performed for each of those graphs. We sampled $100$ trajectories corresponding to a time-window of $5$ units each (roughly $20$ to $30$ transitions). We chose a uniform prior over the graphs and again the uniform prior from the parameter inference above for those. For each of the trajectories containing the $100$ time-windows, we calculated the marginal likelihood \eqref{eq:postgraph} of the graph via sample-based numerical integration employing a Romberg integration and performed a posterior update for the respective survival time $\boldsymbol{\phi}$ and jump parameters $\boldsymbol{\theta}$. Using the calculated marginal likelihood, we did a posterior update for the graph. We saved the result and proceeded in the same way with the other $99$ trajectories until we had $100$ subsequent posterior updates for parameters and graph. The posterior updates for the parameters were dumped afterwards. This has been carried out with all the $500$ graphs, so that we sampled $100$ posterior distributions for each multiple of $5$ units of time over the random graphs. For each of the graph posteriors, we gathered the marginal probabilities of the presence of an edge in a weighted adjacency matrix. Using the true graph and the matrices, we then calculated the AUROC and AUPR classification scores. For the structure inference for CTBNs we followed the original paper \cite{Nodelman2003}. We can see in fig. \ref{fig:inference}, that traditional CTBN's performed surprisingly well when confronted with samples from the Gamma CTBN, but ultimately fail to classify the true graph in both scenarios for the given amount of trajectories. The augmented CTBN, however, converged to the true graph in both scenarios.

\subsubsection{Effects of Shape Parameters}

Since both Weibull and Gamma distributions contain the exponential distribution as a special case for unit shape parameters, we designed a more controlled structure inference scenario for fixed shape parameter values to study its performance impact on traditional CTBN's. Since this impact has shown to be more pronounced for Weibull distributions (see fig. \ref{fig:inference}), we chose five different fixed shape parameters $k$ ranging from $1$ to $9$, increasingly distinguishing the result from an exponential distribution. We then conducted structure inference on $100$ otherwise random graphs, as we did in section \ref{structureinference}, and calculated posterior updates for $50$ trajectories each. In fig. \ref{fig:grn}, we plot the median AUROC classification score along with the $0.2$ and $0.8$ quantiles over the chosen shape parameters for the inferred graphs. It is visible that at larger values, the scores reached by the traditional CTBN become increasingly worse compared to the ones for Weibull CTBN's. The small difference at shape $k=1$, where all distributions are exponential, can be explained by the different prior distributions for the parameters. Nonetheless, it is notable, that the Weibull CTBN shows no disadvantage to the traditional CTBN when confronted with traditional data. The fluctuations in the median can be accounted to numerical inaccuracies and the limited amount of samples. The equal shapes for all involved distributions, however, may also affect the scores for both the Weibull and traditional CTBN's.

\subsection{Gene Regulatory Network}

Finally, we show a scenario on structure inference via time-course data of gene expression through GeneNetWeaver \cite{Schaffter2011}. For this, we selected a gene regulatory network (GRN) from E. Coli consisting of $4$ genes (nodes) and simulated noise-free trajectories at maximum time resolution for a virtual time of 1000 units. We further disabled self-regulation. The exact settings can be found in Appendix B. The values in the trajectories correspond to a relative activation of the gene based on concentrations of species present in the network and therefore continuously range from $0$ to $1$.

As at this stage, we have not formulated our model as a latent model, we need to preprocesses data. At first, we thresholded the obtained data to two levels $0$ and $1$ corresponding to two states per node. To account for the incompleteness, we interpolated the observation points in the trajectories to hold their current value until the next point, at which they instantly change or not, depending on their value. After we now obtained continuous time discrete trajectories, we removed those, which exhibit less than $8$ transitions. The remaining $100$ trajectories were used for structure inference in the same way like explained above in the respective experiment.

We calculated the marginal probabilities of the presence of an edge using the graph posterior and arranged them in a weighted adjacency matrix. We also performed structure inference using a traditional CTBN on the data. In fig. \ref{fig:grn} b) we plot the resulting matrices for both CTBNs. We find that the high-confidence predictions by augmented CTBNs are the correct, while high-confidence predictions of traditional CTBN are incorrect for the given amount of samples.

\section{Conclusion}

In this work, we have presented an augmentation of CTBN's, which aims to inherit all key characteristics of its ancestor, but alleviate its restriction to simple CTMC's, which only allow exponential survival times between state changes. Inference on the model proves to be surprisingly tractable, but the introduction of real-valued clocks raises the models complexity by a significant degree. However, since computational resources are becoming more and more available, it is easier to afford inferring models bearing higher computational cost than it was in the times CTBN's were introduced. Since progress has been made in the field of approximate inference in continuous time discrete state systems with incomplete and noisy evidence recently~\cite{El-Hay2010,Rao2012,Linzner2018,Linzner2019}, more general of such models are not longer considered hopelessly intractable. Especially since PH-distributions are highly over-parameterized and their advantages in the use in CTBN's are restricted to cases where incomplete evidence is present, we aim to present a more general approach to arbitrary survival times in CTBN's here in the light of recent progress made in the field.

\section*{Acknowledgements}
We thank the anonymous reviewers for helpful comments on the previous version of this manuscript.
Nicolai Engelmann and Heinz Koeppl acknowledge support by the European Research Council (ERC) within the CONSYN project, No. 773196.
Dominik Linzner and Heinz Koeppl acknowledge funding by the European Union's Horizon 2020 research and innovation programme (iPC--Pediatric Cure, No. 826121) and (PrECISE, No.  668858). Heinz Koeppl acknowledges support by the Hessian research priority programme LOEWE within the project CompuGene.

\bibliography{main}
\bibliographystyle{icml2020}

\end{document}


\appendix

\section{Proofs and Derivations}

In this section, we provide more detailled derivations
and proofs for the claims in the paper.

\subsection{Global Rates of a CTBN}

A CTBN is (partially) defined via a set of conditional
rates $\lambda_{n}\,:\;\mathcal{X}_{n}\times\mathcal{U}_{n}\rightarrow\mathbb{R}_{\geq0}$,
unique to each process, each local state and parent state. In the
following, we show that the exit rate of the global process
$\lambda\left(\boldsymbol{x}\right)$ decomposes into a sum of these
local rates. 

For this, we notice, that we can express the probabilities of transitioning
within small time steps in terms of rates. For the global process, the probability of
changing in a small window $h$ conditioned on all current states can be formulated as

\begin{align*}
P\left(\bigcup_{n}X_{n}\left(t+h\right)\neq x_{n}\,\biggm|\,\bigcap_{n}X_{n}\left(t\right)=x_{n}\right) & =\sum_{n=1}^{N}P\left(X_{n}\left(t+h\right)\neq x_{n}\,\biggm|\,\bigcap_{m}X_{m}\left(t\right)=x_{m}\right)\qquad\textrm{(asynchronicity)}\\
 & =\sum_{n=1}^{N}P\left(X_{n}\left(t+h\right)\neq x_{n}\,\bigm|\,X_{n}\left(t\right)=x_{n},\,U_{n}\left(t\right)=\boldsymbol{u}_{n}\right)\qquad\textrm{(conditional independence)}\\
 & =\sum_{n=1}^{N}\sum_{x'\neq x_{n}}P\left(X_{n}\left(t+h\right)=x'\,\bigm|\,X_{n}\left(t\right)=x_{n},\,U_{n}\left(t\right)=\boldsymbol{u}_{n}\right)\\
 & =\sum_{n=1}^{N}\lambda_{n}\left(x_{n},\,\boldsymbol{u}_{n}\right)h+o\left(h\right).\qquad\textrm{(def. local exit rates)}
\end{align*}

Subsequently, in the limit $h\rightarrow 0$ we recover $\lambda\left(\boldsymbol{x}\right)$ :

\[
\lambda\left(\boldsymbol{x}\right)=\lim_{h\rightarrow0}\frac{1}{h}\left(\sum_{n=1}^{N}\lambda_{n}\left(x_{n},\,\boldsymbol{u}_{n}\right)h+o\left(h\right)\right)=\sum_{n=1}^{N}\lambda_{n}\left(x_{n},\,\boldsymbol{u}_{n}\right).
\]

\subsection{Transition Probabilities of a CTSMC}

Similarly to CTMC's, we can also assign rates to the associated Markov
process of a CTSMC. The CTSMCs transition-matrix can, as in the case of CTMCs, be expanded in orders of $h$ in terms of instantaneous transition rates employing information of the clock

\begin{gather*}
P\left(X\left(t+h\right)=x',\,T\left(t+h\right)\in\left[0,\,0+h\right)\,|\,X\left(t\right)=x,\,T\left(t\right)\in\left[\tau,\,\tau+h\right)\right)\\
=\lambda\left(x,\,\tau;\,x'\right)h+o\left(h\right).
\end{gather*}

Similarly we can express the probability that no state change occurs

\begin{gather*}
P\left(X\left(t+h\right)=x,\,T\left(t+h\right)\in\left[\tau+h,\,\tau+2h\right)\,|\,X\left(t\right)=x,\,T\left(t\right)\in\left[\tau,\,\tau+h\right)\right)\\
=1-\lambda\left(x,\,\tau\right)h+o\left(h\right).
\end{gather*}

with $\lambda\left(x,\,\tau\right)=\sum_{x'\neq x}\lambda\left(x,\,\tau;\,x'\right)$. Note, that the clock's exit state, does not appear on the rhs's of the equations,
since the clock transitions deterministically under knowledge of the state $X\left(t\right)$, always
satisfying the relation $T\left(t+h\right)-T\left(t\right)\leq h$
for any small time window $h$. The clock's partially deterministic dynamics can be visualized as a
straight line with slope $1$ over time, as long as no state change $X\left(t\right)$ occurs. In case of a state change, however,
the clock resets to zero. This explains the clocks exit state in the first equation, as during the time $h$ a clock reset must have happened.
Since the clocks reachable states are bound by the time window $h$, the state
of the clock must be found within $\left[0,\,0+h\right)$ after $h$
time. In the probability of no state change occurring, the clock keeps rising and is therefore
found to be within $\left[\tau+h,\,\tau+2h\right)$ after $h$ time.
Any multiple state changes within $h$ time become less probable and
finally vanish, once $h$ approaches zero, since they scale with $o\left(h\right)$.

To  obtain the transition probabilities for the state alone, we need to determine
which state is to be chosen, if we know, that a change has happened.
Using basic rules of probability, this leads to

\begin{gather*}
P\left(X\left(t+h\right)=x'\,|\,X\left(t+h\right)\neq x,\,T\left(t+h\right)\in\left[0,\,0+h\right),\,X\left(t\right)=x,\,T\left(t\right)\in\left[\tau,\,\tau+h\right)\right)\\
=\frac{P\left(X\left(t+h\right)=x',\,X\left(t+h\right)\neq x,\,T\left(t+h\right)\in\left[0,\,0+h\right)\,\mid\,X\left(t\right)=x,\,T\left(t\right)\in\left[\tau,\,\tau+h\right)\right)}{P\left(X\left(t+h\right)\neq x,\,T\left(t+h\right)\in\left[0,\,0+h\right)\,\mid\,X\left(t\right)=x,\,T\left(t\right)\in\left[\tau,\,\tau+h\right)\right)}\qquad\textrm{(cond. probability)}\\
=\frac{P\left(X\left(t+h\right)=x',\,T\left(t+h\right)\in\left[0,\,0+h\right)\,\mid\,X\left(t\right)=x,\,T\left(t\right)\in\left[\tau,\,\tau+h\right)\right)}{\lambda\left(x,\,\tau\right)h+o\left(h\right)}\qquad\left(\frac{\textrm{absorption}}{\textrm{def. exit rate}}\right)\\
=\frac{\lambda\left(x,\,\tau;\,x'\right)h+o\left(h\right)}{\lambda\left(x,\,\tau\right)h+o\left(h\right)}\qquad\left(\frac{\textrm{def. transition rates}}{\textrm{}}\right)
\end{gather*}

Taking the limit gives the transition probabilities from state $x$
to $x'$ for a clock that shows $T\left(t\right)=\tau$

\[
\lim_{h\rightarrow0}\left(\frac{\lambda\left(x,\,\tau;\,x'\right)h+o\left(h\right)}{\lambda\left(x,\,\tau\right)h+o\left(h\right)}\right)=\frac{\lambda\left(x,\,\tau;\,x'\right)}{\lambda\left(x,\,\tau\right)}
\]

\subsection{Survival Function and Transition Probabilities of the augmented CTBN}

\subsubsection{Conditional Transition Rates}

Before we can proceed to work out characteristica of the global process
of the augmented CTBN, we need to define the local conditional transition
rates $\lambda_{n}\left(x_{n},\,\tau_{n},\,\boldsymbol{u}_{n};\,x_{n}'\right)$.
These arise naturally in the expressions derived for the global process
under the condition of transition asynchronicity and the restricted
coupling induced by the graph $\mathcal{G}$. We define the rates
$\lambda_{n}\left(x_{n},\,\tau_{n},\,\boldsymbol{u}_{n};\,x_{n}'\right)$
by

\begin{gather*}
P\left(X_{n}\left(t+h\right)=x_{n}',\,T_{n}\left(t+h\right)\in\left[0,\,0+h\right)\,|\,X_{n}\left(t\right)=x_{n},\,T_{n}\left(t\right)\in\left[\tau_{n},\,\tau_{n}+h\right),\,U_{n}\left(t\right)=\boldsymbol{u}_{n}\right)\\
=\lambda_{n}\left(x_{n},\,\tau_{n},\,\boldsymbol{u}_{n};\,x_{n}'\right)h+o\left(h\right)
\end{gather*}

and the local conditional exit rate $\lambda_{n}\left(x_{n},\,\tau_{n},\,\boldsymbol{u}_{n}\right)$
by

\begin{gather*}
\sum_{x_{n}'\neq x_{n}}P\left(X_{n}\left(t+h\right)=x_{n}',\,T_{n}\left(t+h\right)\in\left[0,\,0+h\right)\,|\,X_{n}\left(t\right)=x_{n},\,T_{n}\left(t\right)\in\left[\tau_{n},\,\tau_{n}+h\right),\,U_{n}\left(t\right)=\boldsymbol{u}_{n}\right)\\
=P\left(X_{n}\left(t+h\right)\neq x_{n},\,T_{n}\left(t+h\right)\in\left[0,\,0+h\right)\,|\,X_{n}\left(t\right)=x_{n},\,T_{n}\left(t\right)\in\left[\tau_{n},\,\tau_{n}+h\right),\,U_{n}\left(t\right)=\boldsymbol{u}_{n}\right)\\
=\lambda_{n}\left(x_{n},\,\tau_{n},\,\boldsymbol{u}_{n}\right)h+o\left(h\right)
\end{gather*}

from which we can conclude, that $\lambda_{n}\left(x_{n},\,\tau_{n},\,\boldsymbol{u}_{n}\right)=\sum_{x_{n}'\neq x_{n}}\lambda_{n}\left(x_{n},\,\tau_{n},\,\boldsymbol{u}_{n};\,x_{n}'\right)$
holds in the limit $h\rightarrow0$.

\subsubsection{Global Survival Time}

In order to determine the global survival function and density of the augmented
CTBN, we need to find an expression of the global exit rate. Here, the procedure above
from CTBN's is helpful in combination with the definition
of the transition rates for CTSMC's. The probability, that
a state change of the global process occurs, meaning any of the processes
change and the respective clock resets, is

\begin{gather*}
P\left(\bigcup_{n}X_{n}\left(t+h\right)\neq x_{n},\,T_{n}\left(t+h\right)\in\left[0,\,0+h\right)\,\biggm|\,\bigcap_{n}X_{n}\left(t\right)=x_{n},\,T_{n}\left(t\right)\in\left[\tau_{n},\,\tau_{n}+h\right)\right)\\
=\sum_{n=1}^{N}P\left(X_{n}\left(t+h\right)\neq x_{n},\,T_{n}\left(t+h\right)\in\left[0,\,0+h\right)\,\biggm|\,\bigcap_{m}X_{m}\left(t\right)=x_{m},\,T_{m}\left(t\right)\in\left[\tau_{n},\,\tau_{n}+h\right)\right)\qquad\textrm{(asynchronicity)}\\
=\sum_{n=1}^{N}P\left(X_{n}\left(t+h\right)\neq x_{n},\,T_{n}\left(t+h\right)\in\left[0,\,0+h\right)\,|\,X_{n}\left(t\right)=x_{n},\,T_{n}\left(t\right)\in\left[\tau_{n},\,\tau_{n}+h\right),\,U_{n}\left(t\right)=\boldsymbol{u}_{n}\right)\qquad\textrm{(conditional independence)}\\
=\sum_{n=1}^{N}\lambda_{n}\left(x_{n},\,\tau_{n},\,\boldsymbol{u}_{n}\right)h+o\left(h\right).\qquad\textrm{(def. local exit rates)}
\end{gather*}

Since this covers any process in the network, we can take the limit
to obtain

\[
\lambda\left(\boldsymbol{x},\,\boldsymbol{\tau}\right)=\lim_{h\rightarrow0}\frac{1}{h}\left(\sum_{n=1}^{N}\lambda_{n}\left(x_{n},\,\tau_{n},\,\boldsymbol{u}_{n}\right)h+o\left(h\right)\right)=\sum_{n=1}^{N}\lambda_{n}\left(x_{n},\,\tau_{n},\,\boldsymbol{u}_{n}\right).
\]

Consequently, the complementary probability that no change occurs reads

\begin{gather*}
P\left(\bigcap_{n}X_{n}\left(t+h\right)=x_{n},\,T_{n}\left(t+h\right)\in\left[\tau_{n}+h,\,\tau_{n}+2h\right)\,\biggm|\,\bigcap_{n}X_{n}\left(t\right)=x_{n},\,T_{n}\left(t\right)\in\left[\tau_{n},\,\tau_{n}+h\right)\right)\\
=1-P\left(\bigcup_{n}X_{n}\left(t+h\right)\neq x_{n},\,T_{n}\left(t+h\right)\in\left[0,\,0+h\right)\,\biggm|\,\bigcap_{n}X_{n}\left(t\right)=x_{n},\,T_{n}\left(t\right)\in\left[\tau_{n},\,\tau_{n}+h\right)\right)\\
=1-\sum_{n=1}^{N}\lambda_{n}\left(x_{n},\,\tau_{n},\,\boldsymbol{u}_{n}\right)h+o\left(h\right).
\end{gather*}

We now formulate the path integral corresponding to the global survival function.
For this, we ask for the probability of all processes keeping their
state for $K$ time windows of length $h$ and then perform the continuous time limit. Using the shorthand for this limit $\left(E\equiv\left\{ h\rightarrow0,\,K\rightarrow\infty,\,Kh=s\right\} \right)$ we can write:

\begin{gather}
\label{eq:globsurv}
\Lambda\left(s\,|\,\boldsymbol{\tau},\,\boldsymbol{x}\right)\nonumber \\
=\lim_{E}\prod_{k=0}^{K}P\left(\bigcap_{n}X_{n}\left(t+h\right)=x_{n},\,T_{n}\left(t+h\right)\in\left[\tau+\left(k+1\right)h,\,\tau+\left(k+2\right)h\right)\,\biggm|\,\bigcap_{n}X_{n}\left(t\right)=x_{n},\,T_{n}\left(t\right)\in\left[\tau_{n}+kh,\,\tau_{n}+\left(k+1\right)h\right)\right)\nonumber \\
=\lim_{E}\prod_{k=0}^{K}\left(1-\sum_{n}\lambda_{n}\left(x_{n},\,\tau_{n}+kh,\,\boldsymbol{u}_{n}\right)h\right)+o\left(h\right)=\prod_{0}^{s}\left(1-\sum_{n}\lambda_{n}\left(x_{n},\,\tau_{n}+\sigma,\,\boldsymbol{u}_{n}\right)\mathrm{d}\sigma\right)\qquad\textrm{(Volterra product integral)}\nonumber \\
=\exp\left(-\int_{0}^{s}\mathrm{d}\sigma\,\sum_{n}\lambda_{n}\left(x_{n},\,\tau_{n}+\sigma,\,\boldsymbol{u}_{n}\right)\right)=\exp\left(-\sum_{n}\int_{\tau_{n}}^{s+\tau_{n}}\mathrm{d}\sigma\,\lambda_{n}\left(x_{n},\,\sigma,\,\boldsymbol{u}_{n}\right)\right)\nonumber \\
=\exp\left(-\sum_{n}\int_{0}^{s+\tau_{n}}\mathrm{d}\sigma\,\lambda_{n}\left(x_{n},\,\sigma,\,\boldsymbol{u}_{n}\right)\right)\exp\left(\sum_{n}\int_{0}^{\tau_{n}}\mathrm{d}\sigma\,\lambda_{n}\left(x_{n},\,\sigma,\,\boldsymbol{u}_{n}\right)\right)\nonumber\\
=\prod_{n}\frac{\Lambda_{n}\left(s+\tau_{n}\,|\,x_{n},\,\boldsymbol{u}_{n}\right)}{\Lambda_{n}\left(\tau_{n}\,|\,x_{n},\,\boldsymbol{u}_{n}\right)},\nonumber 
\end{gather}

with having identified $\Lambda_{n}\left(\tau_{n}\,|\,x_{n},\,\boldsymbol{u}_{n}\right)\equiv\exp\left(-\int_{0}^{\tau_{n}}\mathrm{d}\sigma\,\lambda_{n}\left(x_{n},\,\sigma,\,\boldsymbol{u}_{n}\right)\right)$
as the local survival functions. Noticing $F\left(s\,|\,\boldsymbol{\tau},\,\boldsymbol{x}\right)=1-\Lambda\left(s\,|\,\boldsymbol{\tau},\,\boldsymbol{x}\right)$,
we can further recover the density of the global survival time
from the main paper. To do so, we differentiate the global survival time c.d.f.

\begin{align*}
\frac{\mathrm{d}F\left(s\,|\,\boldsymbol{\tau},\,\boldsymbol{x}\right)}{\mathrm{d}s} & =-\frac{\mathrm{d}\Lambda\left(s\,|\,\boldsymbol{\tau},\,\boldsymbol{x}\right)}{\mathrm{d}s}\\
 & \propto\sum_{n}\lambda_{n}\left(x_{n},\,s+\tau_{n},\,\boldsymbol{u}_{n}\right)\exp\left(-\sum_{m}\int_{0}^{s+\tau_{m}}\mathrm{d}\sigma\,\lambda_{m}\left(x_{m},\,\sigma,\,\boldsymbol{u}_{m}\right)\right)\\
 & =\sum_{n}\lambda_{n}\left(x_{n},\,s+\tau_{n},\,\boldsymbol{u}_{n}\right)\prod_{m}\exp\left(-\int_{0}^{s+\tau_{m}}\mathrm{d}\sigma\,\lambda_{m}\left(x_{m},\,\sigma,\,\boldsymbol{u}_{m}\right)\right)\\
 & =\sum_{n}\frac{\mathrm{d}F_{n}\left(s+\tau_{n}\,|\,x_{n},\,\boldsymbol{u}_{n}\right)}{\mathrm{d}s}\prod_{m\neq n}\Lambda_{m}\left(s+\tau_{m}\,|\,x_{m},\,\boldsymbol{u}_{m}\right).
\end{align*}

Note, that we do not need to take care of truncations since they are independent of $s$ and behave like a constant during
differentiation. Reinserting the constant then gives the density of
the global survival time

\begin{align*}
\frac{\mathrm{d}F\left(s\,|\,\boldsymbol{\tau},\,\boldsymbol{x}\right)}{\mathrm{d}s}=\sum_{n}\frac{\mathrm{d}F_{n}\left(s+\tau_{n}\,|\,x_{n},\,\boldsymbol{u}_{n}\right)}{\mathrm{d}s}\frac{1}{\Lambda_{n}\left(\tau_{n}\,|\,x_{n},\,\boldsymbol{u}_{n}\right)}\prod_{m\neq n}\frac{\Lambda_{m}\left(s+\tau_{m}\,|\,x_{m},\,\boldsymbol{u}_{m}\right)}{\Lambda_{m}\left(\tau_{m}\,|\,x_{m},\,\boldsymbol{u}_{m}\right)}.
\end{align*}

\subsubsection{Survival Time Parameterization}

As we derived an expression of the global survival time in terms of local survival functions $\Lambda_{n}\left(s\,|\,x_{n},\,\boldsymbol{u}_{n}\right)$, we are now able to parametrize the local rates of the augmented CTBN. Where in traditional CTBN's, we only have to choose constant rates, in our case, we can assign them arbitrary functions. In this work, choose a desired parametric local survival time distribution and calculate the rates via the known relation

\begin{align*}
\frac{\mathrm{d}\Lambda_{n}\left(s\,|\,x_{n},\,\boldsymbol{u}_{n}\right)}{\mathrm{d}s}&=-\lambda_{n}\left(x_{n},\,s,\,\boldsymbol{u}_{n}\right)\exp\left(-\int_{0}^{s}\mathrm{d}\sigma\,\lambda_{n}\left(x_{n},\,\sigma,\,\boldsymbol{u}_{n}\right)\right)\\
&=-\lambda_{n}\left(x_{n},\,s,\,\boldsymbol{u}_{n}\right)\Lambda_{n}\left(s\,|\,x_{n},\,\boldsymbol{u}_{n}\right)\\\\
\lambda_{n}\left(x_{n},\,s,\,\boldsymbol{u}_{n}\right)&=-\frac{\mathrm{d}\Lambda_{n}\left(s\,|\,x_{n},\,\boldsymbol{u}_{n}\right)}{\mathrm{d}s}\frac{1}{\Lambda_{n}\left(s\,|\,x_{n},\,\boldsymbol{u}_{n}\right)}.
\end{align*}

To give an example, we could choose a Weibull distribution giving $\Lambda_{n}\left(s\,|\,x_{n},\,\boldsymbol{u}_{n}\right)=b k s^{k-1}\exp\left(-b s^k\right)$ with parameters $b$ for the rate and $k$ for the shape. Giving the above equation, we can then calculate $\lambda_{n}\left(x_{n},\,s,\,\boldsymbol{u}_{n}\right)=b k s^{k-1}$. By assigning individual tuples $\left(b,\,k\right)$ to all combinations $n$, $x_n$ and $\boldsymbol{u}_n$, we can construct CTBN's with purely Weibull instead of exponential distributions. 

\subsubsection{Transition Probabilities}

Proceeding from here, we obtain the transition probabilities of the
global process in terms of local rates, using a similar derivation as for CTSMCs

\begin{gather*}
P\left(X_{n}\left(t+h\right)=x_{n}'\,\biggm|\,\bigcup_{m}X_{m}\left(t+h\right)\neq x_{m},\,T_{m}\left(t+h\right)\in\left[0,\,0+h\right),\,\bigcap_{m}X_{m}\left(t\right)=x,\,T_{m}\left(t\right)\in\left[\tau_{m},\,\tau_{m}+h\right)\right)\\
=\frac{P\left(X_{n}\left(t+h\right)=x_{n}',\,\bigcup_{m}X_{m}\left(t+h\right)\neq x_{m},\,T_{m}\left(t+h\right)\in\left[0,\,0+h\right)\,\mid\,\bigcap_{m}X_{m}\left(t\right)=x,\,T_{m}\left(t\right)\in\left[\tau_{m},\,\tau_{m}+h\right)\right)}{P\left(\bigcup_{m}X_{m}\left(t+h\right)\neq x_{m},\,T_{m}\left(t+h\right)\in\left[0,\,0+h\right)\,\mid\,\bigcap_{m}X_{m}\left(t\right)=x,\,T_{m}\left(t\right)\in\left[\tau_{m},\,\tau_{m}+h\right)\right)}\\
=\frac{P\left(\bigcup_{m}X_{n}\left(t+h\right)=x_{n}',\,X_{m}\left(t+h\right)\neq x_{m},\,T_{m}\left(t+h\right)\in\left[0,\,0+h\right)\,\mid\,\bigcap_{m}X_{m}\left(t\right)=x,\,T_{m}\left(t\right)\in\left[\tau_{m},\,\tau_{m}+h\right)\right)}{\sum_{n=1}^{N}\lambda_{n}\left(x_{n},\,\tau_{n},\,\boldsymbol{u}_{n}\right)h+o\left(h\right)}\quad\left(\frac{\textrm{distributiveness}}{\textrm{global exit rate}}\right)\\
=\frac{P\left(X_{n}\left(t+h\right)=x_{n}',\,T_{n}\left(t+h\right)\in\left[0,\,0+h\right)\,\mid\,\bigcap_{m}X_{m}\left(t\right)=x,\,T_{m}\left(t\right)\in\left[\tau_{m},\,\tau_{m}+h\right)\right)}{\sum_{n=1}^{N}\lambda_{n}\left(x_{n},\,\tau_{n},\,\boldsymbol{u}_{n}\right)h+o\left(h\right)}\quad\left(\frac{\textrm{asynchronicity \& absorption}}{\textrm{}}\right)\\
=\frac{\lambda_{n}\left(x_{n},\,\tau_{n},\,\boldsymbol{u}_{n};\,x_{n}'\right)h+o\left(h\right)}{\sum_{n=1}^{N}\lambda_{n}\left(x_{n},\,\tau_{n},\,\boldsymbol{u}_{n}\right)h+o\left(h\right)}\quad\qquad\left(\frac{\textrm{local transition rates}}{\textrm{}}\right).
\end{gather*}

Finally, taking the limit gives us the instantaneous transition probabilities of the global
process given the clocks state $\boldsymbol{T}\left(t\right)=\boldsymbol{\tau}$

\[
\lim_{h\rightarrow0}\left(\frac{\lambda_{n}\left(x_{n},\,\tau_{n},\,\boldsymbol{u}_{n};\,x_{n}'\right)h+o\left(h\right)}{\sum_{n=1}^{N}\lambda_{n}\left(x_{n},\,\tau_{n},\,\boldsymbol{u}_{n}\right)h+o\left(h\right)}\right)=\frac{\lambda_{n}\left(x_{n},\,\tau_{n},\,\boldsymbol{u}_{n};\,x_{n}'\right)}{\lambda\left(\boldsymbol{x},\,\boldsymbol{\tau}\right)}.
\]

\subsection{Derivation of the Path Measure}

After having obtained the probabilities belonging to the generative process, it is straight forward to give the path density for a single interval. Assume, the associated
Markov process of the global process is in state $\left(\boldsymbol{x},\,\boldsymbol{\tau}\right)$
at the beginning of the interval. We keep $\boldsymbol{x}$
for a time exactly $s$ and then, the $n$-th process transitions
from its state $x_{n}$ to $x_{n}'$. The density for this event is
then 

\begin{align*}
p\left(\boldsymbol{x}',\,\boldsymbol{\tau}'\,|\,\boldsymbol{x},\,\boldsymbol{\tau},\,\boldsymbol{\theta}\right)&=\frac{\lambda_{n}\left(x_{n},\,\tau_{n}+s,\,\boldsymbol{u}_{n};\,x_{n}'\right)}{\sum_{n=1}^{N}\lambda_{n}\left(x_{n},\,\tau_{n}+s,\,\boldsymbol{u}_{n}\right)}\frac{\mathrm{d}F\left(s\,|\,\boldsymbol{\tau},\,\boldsymbol{x}\right)}{\mathrm{d}s}\\
&\propto\frac{\lambda_{n}\left(x_{n},\,\tau_{n}+s,\,\boldsymbol{u}_{n};\,x_{n}'\right)}{\sum_{n=1}^{N}\lambda_{n}\left(x_{n},\,s+\tau_{n},\,\boldsymbol{u}_{n}\right)}\sum_{n}\lambda_{n}\left(x_{n},\,s+\tau_{n},\,\boldsymbol{u}_{n}\right)\prod_{m}\exp\left(-\int_{0}^{s+\tau_{m}}\mathrm{d}\sigma\,\lambda_{m}\left(x_{m},\,\sigma,\,\boldsymbol{u}_{m}\right)\right)\\
&=\lambda_{n}\left(x_{n},\,\tau_{n}+s,\,\boldsymbol{u}_{n};\,x_{n}'\right)\exp\left(-\int_{0}^{s+\tau_{n}}\mathrm{d}\sigma\,\lambda_{n}\left(x_{n},\,\sigma,\,\boldsymbol{u}_{n}\right)\right)\prod_{m\neq n}\Lambda_{m}\left(s+\tau_{m}\,|\,x_{m},\,\boldsymbol{u}_{m}\right),
\end{align*}
with 
\begin{align}\label{eq:update}
\boldsymbol{x}' & =\left(x_{1},\,\dots,\,x_{n-1},\,x_{n}',\,x_{n+1},\,\dots,\,x_{N}\right),\\
\boldsymbol{\tau}' & =\left(\tau_{1}+s,\,\dots,\,\tau_{n-1}+s,\,0,\,\tau_{n+1}+s,\,\dots,\,\tau_{N}+s\right).\nonumber
\end{align}

If we further restrict ourselves to processes obeying time-direction
independence, we can express the density in terms of transition probabilities $\lambda_{n}\left(x_{n},\,\tau_{n}+s,\,\boldsymbol{u}_{n};\,x_{n}'\right)$
$=\theta_{n}\left(x_{n}'\,\mid\,x_{n},\,\boldsymbol{u}_{n}\right)$
$\lambda_{n}\left(x_{n},\,\tau_{n}+s,\,\boldsymbol{u}_{n}\right)$
and obtain after reinserting the constant term representing the truncations

\begin{align*}
p\left(\boldsymbol{x}',\,\boldsymbol{\tau}'\,|\,\boldsymbol{x},\,\boldsymbol{\tau},\,\boldsymbol{\theta}\right)&=\theta_{n}\left(x_{n}'\,\mid\,x_{n},\,\boldsymbol{u}_{n}\right)\lambda_{n}\left(x_{n},\,\tau_{n}+s,\,\boldsymbol{u}_{n}\right)\frac{\exp\left(-\int_{0}^{s+\tau_{n}}\mathrm{d}\sigma\,\lambda_{n}\left(x_{n},\,\sigma,\,\boldsymbol{u}_{n}\right)\right)}{\Lambda_{n}\left(\tau_{n}\,|\,x_{n},\,\boldsymbol{u}_{n}\right)}\prod_{m\neq n}\frac{\Lambda_{m}\left(s+\tau_{m}\,|\,x_{m},\,\boldsymbol{u}_{m}\right)}{\Lambda_{m}\left(\tau_{m}\,|\,x_{m},\,\boldsymbol{u}_{m}\right)}\\
&=\theta_{n}\left(x_{n}'\,\mid\,x_{n},\,\boldsymbol{u}_{n}\right)\frac{\mathrm{d}F_{n}\left(s+\tau_{n}\,|\,x_{n},\,\boldsymbol{u}_{n}\right)}{\mathrm{d}s}\frac{1}{\Lambda_{n}\left(\tau_{n}\,|\,x_{n},\,\boldsymbol{u}_{n}\right)}\prod_{m\neq n}\frac{\Lambda_{m}\left(s+\tau_{m}\,|\,x_{m},\,\boldsymbol{u}_{m}\right)}{\Lambda_{m}\left(\tau_{m}\,|\,x_{m},\,\boldsymbol{u}_{m}\right)}
\end{align*}

We observe, that one part in this expression depends on $\boldsymbol{x}$ and $\boldsymbol{x}'$ and one
on $\boldsymbol{x}$, $\boldsymbol{\tau}$ and $s$. The first is
the transition probability associated with the embedded Markov chains
of the single processes with time-direction independence. To infer
those, we do not need any timing information. On the other hand, to
infer the second expression related to the survival times, we do need
timing information.

\subsection{Likelihood in a Gamma CTBN}

In order to derive the likelihood function of a single survival time parameter
tuple $\left(\alpha_{n},\,\beta_{n}\right)$ associated with the $n$-th
process, its state $x$ and parent state $\boldsymbol{u}$ in a Gamma CTBN, we need to build the product of the respective censored,
non-censored and truncated factors of the global likelihood. First consider the sets $\mathbf{X}$ and $\mathbf{T}$ from the main text. The set $\mathbf{X}$ contains samples of the global state at the beginning of each interval. On the other hand, $\mathbf{T}$ contains samples of all clock values at the beginning of each interval. Let $\mathbf{X}_{n}\subset\mathbf{X}$
and $\mathbf{T}_{n}\subset\mathbf{T}$ be the subset of samples
of states and clock values relevant to the $n$-th process. This means,
that $\mathbf{X}_{n}$ only contains the sample values from $\mathbf{X}$ corresponding to the $n$-th process and its parents. The $\mathbf{T}_{n}$ only contains the sample values of the $n$-th clock from $\mathbf{T}$. Further, there are no consecutive repetitions in $\mathbf{X}_{n}$
meaning, that one element in $\mathbf{X}_{n}$ is not associated
with two consecutive intervals. Since we assumed all parameter sets
for individual $x$ and $\boldsymbol{u}$ to be independent, samples
contained in $\mathbf{X}_{n}$ and $\mathbf{T}_{n}$ appear
as constants and are normalized out when building the posterior density
for $\left(\alpha_{n},\,\beta_{n}\right)$. We introduce the following sets:
 $S_{f}\equiv\left\{ s_{f,1},\,s_{f,2},\,\dots\right\} $
consists of all clock samples $s_{f,m}\in\mathbf{T}_{n}$ at
which the $n$-th process transitions, $S_{c}$ consists of
all clock samples, where a parent has changed and $S_{t}$, which consists
of the clock values at the beginning of the intervals. Consider an element
$\left(x,\,\boldsymbol{u}\right)\in\mathbf{X}_{n}$ associated
with the $i$-th interval and another element $\left(x',\,\boldsymbol{u}'\right)\in\mathbf{X}_{n}$
associated with the $\left(i+1\right)$-th interval. If now $x'\neq x$,
then the clock sample $s\in\mathbf{T}_{n}$ associated with the
$\left(i+1\right)$-th interval is an element of $S_{f}$. Otherwise
($x'=x$), $s$ is an element of $S_{c}$. Additionally, $s_{t}\in\mathbf{t}_{n}$
associated with the $i$-th interval is an element of $S_{t}$ in
both cases.

Then, the likelihood
is a product of terms of the Gamma p.d.f. $\frac{\beta_{n}}{\Gamma\left(\alpha_{n}\right)}\left(\beta_{n}s_{f,m}\right)^{\alpha_{n}-1}\exp\left(-\beta_{n}s_{f,m}\right)$
for each $s_{f,m}\in S_{f}$, of the Gamma survival function $\frac{\Gamma\left(\alpha_{n},\,\beta_{n}s_{c,m}\right)}{\Gamma\left(\alpha_{n}\right)}$for
each element in $S_{c}$, and the reciprocal $\frac{\Gamma\left(\alpha_{n}\right)}{\Gamma\left(\alpha_{n},\,\beta_{n}s_{t,m}\right)}$
for each element in $S_{t}$. Building the product, we then obtain
for the posterior update of a full trajectory of the global process
associated with the parameters $\left(\alpha_{n},\,\beta_{n}\right)$

\begin{align*}
 & p\left(\alpha_{n},\,\beta_{n}\,\mid\,\mathbf{X}_{n},\,\mathbf{T}_{n}\right)\\
 & \quad\propto\beta_{n}^{\left|S_{f}\right|}\left(\prod_{m=1}^{\left|S_{f}\right|}\beta_{n}s_{f,m}\right)^{\alpha_{n}-1}\exp\left(-\beta_{n}\sum_{m=1}^{\left|S_{f}\right|}s_{f,m}\right)\\
 & \qquad\frac{\Gamma\left(\alpha_{n}\right)^{\left|S_{t}\right|}\prod_{m=1}^{\left|S_{c}\right|}\Gamma\left(\alpha_{n},\,\beta_{n}s_{c,m}\right)}{\Gamma\left(\alpha_{n}\right)^{\left|S_{f}\right|}\Gamma\left(\alpha_{n}\right)^{\left|S_{c}\right|}\prod_{m=1}^{\left|S_{t}\right|}\Gamma\left(\alpha_{n},\,\beta_{n}s_{t,m}\right)}p\left(\alpha_{n},\,\beta_{n}\right)
\end{align*}

with an arbitrary prior distribution $p\left(\alpha_{n},\,\beta_{n}\right)$.
The Weibull posterior update from the main text is constructed in
a similar way.

\subsection{Inference of Rayleigh CTBN's}

Like we have shown, that the prior distribution for a whole CTBN can be given as a product of local prior distributions, the Rayleigh CTBN has a conjugate prior in the form of a product of inverse Gamma distributions under its typical parametrization $\phi_n\left(x,\,\boldsymbol{u}\right)\equiv \sigma_n\left(x,\,\boldsymbol{u}\right)^2$. We can give this by

\begin{align*}
    p\left(\boldsymbol{\phi}\right)\;
 & \propto \; \prod_n \prod_{\boldsymbol{u} \in \mathcal{U}_n}\prod_{x \in \mathcal{X}_n}\frac{\beta_n\left(x,\,\boldsymbol{u}\right)^{\alpha_n\left(x,\,\boldsymbol{u}\right)}}{\phi_n\left(x,\,\boldsymbol{u}\right)^{\alpha_n\left(x,\,\boldsymbol{u}\right) + 1}}\exp\left(-\frac{\beta_n\left(x,\,\boldsymbol{u}\right)}{\phi_n\left(x,\,\boldsymbol{u}\right)}\right).
\end{align*}

Because we can perform a normalization after multiplying with the likelihood, we can effectively ignore the constant factor $\beta_n\left(x,\,\boldsymbol{u}\right)^{\alpha_n\left(x,\,\boldsymbol{u}\right)}$ and obtain for the posterior update

\begin{align*}
    \alpha_n\left(x,\,\boldsymbol{u}\right) &\rightarrow \alpha_n\left(x,\,\boldsymbol{u}\right) + \left|S_f\right|\\
    \beta_n\left(x,\,\boldsymbol{u}\right) &\rightarrow \beta_n\left(x,\,\boldsymbol{u}\right) + \frac{1}{2}\left(\sum_{m\in S_f \cap S_c} \tau_m^2 - \sum_{m\in S_t} \tau_m^2\right).
\end{align*}

where we can immediately spot the sufficient statistics

\begin{align*}
    T_\beta\left(n,\,x,\,\boldsymbol{u}\right) &= \frac{1}{2}\left(\sum_{m\in S_f \cap S_c} \tau_m^2 - \sum_{m\in S_t} \tau_m^2\right)\\
    T_\alpha\left(n,\,x,\,\boldsymbol{u}\right) &= \left|S_f\right|.
\end{align*}

Again exploiting the normalization of the resulting expression, by comparison of the product of likelihood and prior with the normalized inverse Gamma distribution, we can formulate the marginal likelihood for structure inference by

\begin{align*}
    \int_{\boldsymbol{\Phi}}\,\mathrm{d}\boldsymbol{\phi}\;p\left(\mathbf{T}\,|\,\mathbf{X},\,\boldsymbol{\phi}\right)p\left(\boldsymbol{\phi}\,|\,\mathcal{G}\right) = \prod_n \prod_{\boldsymbol{u} \in \mathcal{U}_n}\prod_{x \in \mathcal{X}_n}\frac{\left(\prod_{m\in S_f\left(n,x,\boldsymbol{u}\right)}\tau_m\right)\beta_n\left(x,\,\boldsymbol{u}\right)^{\alpha_n\left(x,\,\boldsymbol{u}\right)}\Gamma\left(\alpha_n\left(x,\,\boldsymbol{u}\right) + T_\alpha\left(n,\,x,\,\boldsymbol{u}\right)\right)}{\Gamma\left(\alpha_n\left(x,\,\boldsymbol{u}\right)\right)\left(\beta_n\left(x,\,\boldsymbol{u}\right) + T_\beta\left(n,\,x,\,\boldsymbol{u}\right)\right)^{\alpha_n\left(x,\,\boldsymbol{u}\right) + T_\alpha\left(n,\,x,\,\boldsymbol{u}\right)}}
\end{align*}

which can be efficiently calculated using Stirling's approximation of the gamma function.

\section{Configuration of GeneNetWeaver}

In this section, we provide the exact settings for GeneNetWeaver from
the GRN-scenario. As mentioned in the main text, we chose the maximum time-resolution
and simulated until $10^{3}$ units of time. Further, since we perform inference on the latent model directly, we chose an ODE-based simulation without additional noise. We chose "Time series as in DREAM4" and then the following additional settings

\begin{center}
\begin{tabular}{l|r}
    Duration of each time series (t\_max)   &  $1\,000$\\
    Number of measured points...                    &  $1\,001$\\
    Perturbations for multifactorial...             &  Generate new\\
    Noise added after... (measurement error)        &  None\\
\end{tabular}
\end{center}

The resulting time-series' were then preprocessed like mentioned in the main text and the resulting trajectories were then used to train the latent model with a Gamma survival time parametrization.

\section{Simulation Algorithm}

The proposed algorithm for the simulation of the augmented CTBN corresponds to the Gillespie algrorithm for the augmented CTBN and is given below. To draw from a minimum of truncated distributions, we draw from multiple truncated distributions and store the minimum. Additionally, we can draw from the truncated distributions by a simple rejection sampling procedure, continuously drawing samples from the original distribution until the outcome is larger than the value of the truncation. In our experiments, no case has occurred, where this lead to significant increases in runtime. However, care must be taken when single processes exhibit an extreme dynamic range. This can potentially necessitate large amounts of draws after unfavorable parent changes.
\newline

\begin{algorithm}[H]
\SetAlgoLined
\KwResult{A sample trajectory of the augmented CTBN from $0$ to $T$}
 Require $\theta$, $\phi$, $\boldsymbol{x}^{\left(0\right)}$, $\boldsymbol{\tau}^{\left(0\right)}$ (clocks may be all zero)\;
 Set $c \leftarrow 0$\;
 Set $t^{(0)} \leftarrow 0$\;
 \While{$t$ < $T$}{
  Draw $s$ from eq. (3) (main text)\;
  Draw $n$ and $x_n'$ from the categoricals in (5) (main text)\;
  Set $t^{(c+1)} \leftarrow t^{(c)} + s$\;
  \eIf{$t^{(c+1)} > T$}{
   break\;
   }{
   Update $\boldsymbol{x}^{\left(c+1\right)}$ and $\boldsymbol{\tau}^{\left(c+1\right)}$ from $\boldsymbol{x}^{\left(c\right)}$ and $\boldsymbol{\tau}^{\left(c\right)}$ according to \eqref{eq:update} (here in appendix)\;
   Set $c \leftarrow c + 1$\;
  }
  return $\mathbf{X}=\left\{ \boldsymbol{x}^{(0)},\,\boldsymbol{x}^{(1)},\,\dots,\,\boldsymbol{x}^{(c)}\right\}$, $\mathbf{T}=\left\{ \boldsymbol{\tau}^{(0)},\,\boldsymbol{\tau}^{(1)},\,\dots,\,\boldsymbol{\tau}^{(c)}\right\}$ and $\boldsymbol{t}=\left\{ t^{(0)},\,t^{(1)},\,\dots,\,t^{(c)}\right\}$\;
 }
 \caption{Gillespie algorithm for the augmented CTBN}
\end{algorithm}
\vspace{0.5cm}

$\mathbf{X}$ then contains the states $\boldsymbol{x}^{(m)}$ in the time-window $\left[t^{(m)}, t^{(m+1)}\right)$. $\mathbf{T}$ contains the clock values at $\boldsymbol{\tau}^{(m)}$ after transition to the new state. Additionally, the value $c$ contains the number of transitions occurred.